\documentclass[runningheads]{llncs}
\usepackage{times}
\usepackage{soul}
\usepackage{url}
\usepackage[hidelinks]{hyperref}
\usepackage{graphicx}
\usepackage{booktabs}
\usepackage{algorithm}
\usepackage{algorithmic}
\usepackage{bbding} %for star symbols
\usepackage{amssymb}
\usepackage{tikz}
\usetikzlibrary{arrows}
\usetikzlibrary{arrows.meta}

\tikzset{every loop/.style={min distance=6mm,in=-11,out=11,looseness=10}}

\usepackage{pgf}
\usetikzlibrary{arrows,automata}
\usetikzlibrary{positioning}

\usetikzlibrary{calc}

\tikzset{
    between/.style args={#1 and #2}{
         at = ($(#1)!0.5!(#2)$)
    }
}

\tikzset{
    state/.style={
           rectangle,
           rounded corners,
           draw=black, very thick,
           minimum height=2em,
           inner sep=2pt,
           text centered,
           },
}
\definecolor{commentRed}{cmyk}{0, 0.90, 0.60, 0}
\definecolor{commentBlue}{cmyk}{0.8, 0, 0.30, 0.1}

\usepackage{hyperref}
\definecolor{royalblue(traditional)}{rgb}{0.0, 0.14, 0.4}
\hypersetup{
    colorlinks=true,       % false: boxed links; true: colored links
    linkcolor=royalblue(traditional),          % color of internal links
    citecolor=royalblue(traditional),       % color of links to bibliography
    urlcolor=royalblue(traditional)
}
\usepackage{thmtools}
\usepackage{thm-restate}
\usepackage{cleveref}

\newenvironment{pproof}{\par\vspace*{0.5ex}\noindent\rm {\bf Proof:} }
{
}

\newcommand{\mods}[1]{[\![#1]\!]}
\newcommand{\bel}[1]{[#1]}
\newcommand{\cbel}[1]{[#1]_{\mathrm{c}}}
\newcommand{\contract}{\div} 

\newcommand{\STQ}{\oplus_{\mathrm{STQ}}}
\newcommand{\TPO}[1]{\mathrm{TPO}(#1)}

\newcommand{\BetaRevR}[1]{$(\beta{#1}^{\ast}_{\scriptscriptstyle \preccurlyeq})$}

\newcommand{\BetaRevPlusR}[1]{$(\beta{#1 +}^{\ast}_{\scriptscriptstyle \preccurlyeq})$}

\newcommand{\CRevR}[1]{$(\mathrm{C}{#1}^{\ast}_{\scriptscriptstyle \preccurlyeq})$}
\newcommand{\CRevS}[1]{$(\mathrm{C}{#1}^{\ast})$}
\newcommand{\CConR}[1]{$(\mathrm{C}{#1}^{\scriptscriptstyle \contract}_{\scriptscriptstyle \preccurlyeq})$}

\newcommand{\CConRevR}[1]{$(\mathrm{C}{#1}^{\scriptscriptstyle \contract/\ast}_{\scriptscriptstyle \preccurlyeq})$}

\newcommand{\IndRevR}{$(\mathrm{P}^{\ast}_{\scriptscriptstyle \preccurlyeq})$}

\newcommand{\KRev}[1]{$(\mathrm{K}{#1}^{\ast})$}

\newcommand{\KCon}[1]{$(\mathrm{K}{#1}^{\scriptscriptstyle \contract})$}

\newcommand{\HI}{$(\mathrm{HI})$}

\newcommand{\LI}{$(\mathrm{LI})$}

\newcommand{\iLIR}{$(\mathrm{iLIRC})$}

\newcommand{\iLIRSem}{$(\mathrm{iLIRC_{\preccurlyeq}})$}
\newcommand{\iLINRSem}{$(\mathrm{iLI\ast_{\preccurlyeq}})$}

\newcommand{\SPU}{$(\mathrm{SPU}_{\preccurlyeq})$}

\newcommand{\WPU}{$(\mathrm{WPU}_{\preccurlyeq})$}

\newcommand{\NLI}{$(\mathrm{NLI}_{\scriptscriptstyle \preccurlyeq})$}

\newcommand{\IIAPriorRev}{$(\mathrm{IIAP}^{\scriptscriptstyle \ast}_{\scriptscriptstyle \preccurlyeq})$}
\newcommand{\IIAInputRev}{$(\mathrm{IIAI}^{\scriptscriptstyle \ast}_{\scriptscriptstyle \preccurlyeq})$}

\newcommand{\NeutralityRev}{$(\mathrm{Neut}^{\scriptscriptstyle \ast}_{\scriptscriptstyle \preccurlyeq})$}
\newcommand{\REDRev}[1]{$(\mathrm{Red}{#1})$}
\newcommand{\Cn}{\mathrm{Cn}}
\newcommand{\C}{\mathrm{C}}
\newcommand{\CRat}{\mathrm{C_{rat}}}

\newcommand{\RID}{$(\mathrm{RID}^\mathrm{c})$} %Rational Identity
\newcommand{\astN}{\ast_{\mathrm{N}}}
\newcommand{\astR}{\ast_{\mathrm{R}}}
\newcommand{\astL}{\ast_{\mathrm{L}}}

\begin{document}

\title{Elementary Iterated Revision and the Levi Identity\thanks{This research was partially supported by the Australian Government through an ARC Future Fellowship (project number FT160100092) awarded to Jake Chandler.}
}

% Multiple author syntax (remove the single-author syntax above and the \iffalse ... \fi here)
% Check the ijcai19-multiauthor.tex file for detailed instructions
%\iffalse
%\author{
%Jake Chandler$^1$
%\and
%%Second Author$^2$\and
%%Third Author$^{2,3}$\And
%Richard Booth$^2$
%\affiliations
%$^1$La Trobe University
%$^2$Cardiff University\\
%\emails
%jake.chandler@cantab.net,
%BoothR2@cardiff.ac.uk
%}

\author{Jake Chandler\inst{1} \and
Richard Booth\inst{2}}
\authorrunning{J. Chandler and R. Booth}
\institute{La Trobe University, Melbourne, Australia \and
Cardiff University, Cardiff, UK
}

\maketitle

\begin{abstract}
Recent work has considered the problem of extending to the case of iterated belief change the so-called  `Harper Identity' (HI), which defines single-shot contraction in terms of single-shot revision. The present paper considers the prospects of providing a similar extension of the Levi Identity (LI), in which the direction of definition runs the other way.  We restrict our attention here to the three classic iterated revision operators--natural, restrained and lexicographic, for which we provide here the first collective characterisation in the literature, under the appellation of `elementary' operators. We consider two prima facie plausible ways of extending (LI). The first proposal involves the use of the rational closure operator to offer a `reductive' account of iterated revision in terms of iterated contraction. The second, which doesn't commit to reductionism, was put forward some years ago by Nayak {\em et al}. We establish that, for elementary revision operators and under mild assumptions regarding contraction, Nayak's proposal is equivalent to a new  set of postulates formalising the claim that contraction by $\neg A$ should be considered to be a kind of `mild' revision by $A$. We then show that these, in turn, under slightly weaker assumptions, jointly amount to the conjunction of a pair of constraints on the extension of (HI) that were recently proposed in the literature. Finally, we consider the consequences of endorsing both suggestions and show that this would yield an identification of rational revision with natural revision. We close the paper by discussing the general prospects for defining iterated revision in terms of iterated contraction.
\end{abstract}

\section{Introduction}

The crucial question of iterated belief change--that is, the question of the rationality constraints that govern the beliefs  resulting from a sequence of changes in view--remains very much a live one. 

In recent work \cite{DBLP:conf/ijcai/BoothC16}, we have studied in some detail the problem of extending, to the iterated case, a principle of single-step change known as the `Harper Identity' (henceforth `\HI') \cite{harper1976rational}. This principle connects single-step contraction and revision, the two main types of change found in the literature, in a manner that allows one to define the former in terms of the latter. We presented a family of extensions of  \HI~characterised by the satisfaction of an intuitive pair of principles and showed how these postulates could be used to translate principles of iterated revision into principles of iterated contraction.

But \HI~also has a well known companion principle which reverses the direction of definition, allowing one to define single-step revision in terms of single-step contraction: the Levi Identity (henceforth `\LI') \cite{levi1977subjunctives}. To date, furthermore, the issue of extending \LI~to the iterated case has barely been discussed.  
Two noteworthy exceptions are the short papers of Nayak et al \cite{DBLP:conf/dagstuhl/NayakGOP05}  and of Konieczny \& Pino P\'erez \cite{10.1007/978-3-319-67582-4_25}. The second paper argues that no reasonable extension of \LI~will enable us to reduce iterated revision to iterated contraction. The first paper introduces a non-reductionist extension of \LI~consonant with this claim. 

The present contribution aims to provide a more comprehensive discussion of the issue, carried out against the backdrop of the aforementioned recent work on \HI. The plan of the paper is as follows. After a preliminary introduction of the formal framework in Section \ref{s:Preliminaries}, we provide, in Section \ref{s:Elementary}, a novel result that is of general interest in itself. We collectively characterise the three classic belief revision operators that are the focus of the paper (natural, restrained and lexicographic) under the appellation of `elementary' operators, showing that they are in fact the {\em only} operators satisfying a particular set of properties. Section \ref{s:iLI} turns to the issue of extending \LI~to the iterated case. We present, in Section \ref{ss:ThreeRed}, an extension of \LI~based on the concept of rational closure, which would result in a reduction of two-step revision to two-step contraction.
% The first two have rather obvious shortcomings. The third, which is based on the operation of rational closure, is noted to be somewhat restrictive when coupled with a plausible candidate for extending the Harper Identity. Indeed, among the elementary iterated revision operators, only natural revision can simultaneously meet both desiderata.
 Section \ref{ss:NLI} then discusses the non-reductive proposal of \cite{DBLP:conf/dagstuhl/NayakGOP05}. We first 
%demonstrate that it proposal is the only viable option among a broader set of possibilities. Further, we 
establish that, for elementary revision operators and under mild assumptions regarding contraction, it is in fact equivalent to a new set of postulates formalising the claim that contraction by $\neg A$ should be considered to be a kind of `mild' revision by $A$. These, in turn, under slightly weaker assumptions, are  proven to jointly amount to the conjunction of the aforementioned constraints on the extension of \HI~that were proposed in \cite{DBLP:conf/ijcai/BoothC16}.  In Section \ref{ss:RCNLI}, we consider the consequences of endorsing both suggestions and show that this would yield an identification of rational revision with natural revision. In Section \ref{s:reducibility}, we briefly discuss the general prospects for defining iterated revision from iterated contraction, critically assessing the central argument of \cite{10.1007/978-3-319-67582-4_25}. We conclude, in Section \ref{s:Conclusions}, with some remaining open questions. 

The proofs of the various propositions and theorems have been relegated to a technical appendix.

\section{Preliminaries} 
\label{s:Preliminaries}

The beliefs of an agent are represented by a {\em belief state} $\Psi$. The latter determines a {\em belief set} $\bel{\Psi}$, a deductively closed set of sentences, drawn from a finitely generated,  propositional, truth-functional language $L$.  The set of classical logical consequences of $\Gamma\subseteq L$ will be denoted by $\mathrm{Cn}(\Gamma)$. The set of propositional worlds or valuations will be denoted by $W$, and the set of models of a given sentence $A$ by $\mods{A}$.

We consider the three classic belief change operations mapping a prior state $\Psi$ and input sentence $A$ in $L$  onto a posterior state.  The operation of {\em revision} $\ast$ returns the posterior state $\Psi \ast A$ that results from an adjustment of $\Psi$ to accommodate the inclusion of $A$, in such a way as to maintain consistency of the resulting belief set when $\neg A\in\bel{\Psi}$. The operation of {\em expansion} $+$ is similar, save that consistency of the resulting beliefs needn't be ensured. Finally, the operation of {\em contraction} $\contract$ returns the posterior state $\Psi \contract A$ that results from an adjustment of $\Psi$ to accommodate the retraction of $A$. 

\subsection{Single-step change}

In terms of single-step change, revision and contraction are assumed to satisfy the postulates of Alchourr\'on, G\"ardenfors and Makinson \cite{alchourron1985logic} (henceforth `AGM'), while the behaviour of expansion is constrained by $\bel{\Psi + A} = \Cn (\bel{\Psi}\cup\{A\})$. AGM ensures a useful order-theoretic representability of the single-shot revision or contraction dispositions of an agent, such that each $\Psi$ is associated with  a total preorder (henceforth `{\em TPO}')  $\preccurlyeq_\Psi$ over $W$, such that $\mods{\bel{\Psi\ast A}} = \min(\preccurlyeq_\Psi, \mods{A})$ (\cite{grove1988two,katsuno1991propositional}). In this context, the AGM postulate of {\em Success} ($A\in [\Psi* A]$) corresponds to the requirement that $\min(\preccurlyeq_{\Psi \ast A}, W) \subseteq \mods{A}$. We denote by TPO($W$) the set of all TPOs over $W$ and shall assume that, for every $\preccurlyeq\in\TPO{W}$, there is a state $\Psi$ such that $\preccurlyeq=\preccurlyeq_{\Psi}$. 

Equivalently, these revision dispositions can be represented by a `{\em conditional belief set}' $\cbel{\Psi}$. This set extends the belief set $\bel{\Psi}$ by further including various `conditional beliefs', expressed by sentences of the form $A \Rightarrow B$, where $\Rightarrow$ is a non-truth-functional conditional connective and $A, B\in L$ (we shall call $L_c$ the language that extends $L$ to include such conditionals). This is achieved by means of the so-called {\em Ramsey Test}, according to which $A \Rightarrow B \in \cbel{\Psi}$ iff $B \in \bel{\Psi \ast A}$. %\footnote{As is well known \cite{Gardenfors1986-GRDBRA}, the AGM postulates, which quantify over sentences in $L$ and sets thereof, cannot be straightforwardly extended to sentences in $L_c$ and sets thereof. We do not assume any such extension here.}  
 In terms of constraints on $\cbel{\Psi}$, AGM notably ensures that its conditional  subset corresponds to a {\em rational consequence relation}, in the sense of \cite{lehmann1992does} (we shall say, in this case, that $\cbel{\Psi}$ is rational). 
%Since the TPO $\preceq_\Psi$ uniquely determines the conditional belief set $\cbel{\Psi}$, we will sometimes denote the latter by  $\cbel{\preceq_\Psi}$. \jake{check that we still need this last convention}

Following convention, we shall call principles couched in terms of belief sets `syntactic', and call `semantic' those principles couched in terms of TPOs, denoting the latter by subscripting the corresponding syntactic principle with `$\preccurlyeq$'. 

The operations $\ast$ and $\contract$ are assumed to be related in the single-shot case by the Levi and Harper identities, namely

\begin{tabbing}
 \=BLAHBL\=\kill
\> \LI \> $\bel{\Psi \ast A} = \mathrm{Cn}(\bel{\Psi \contract \neg A} \cup \{A\})$ \\ 
\>\HI   \> $\bel{\Psi \contract A} =\bel{\Psi} \cap \bel{\Psi \ast  \neg A}$ \\[-0.25em]
\end{tabbing} 
\vspace{-0.75em}

\noindent with single-shot revision determining single-shot expansion via a third identity:

 \begin{tabbing}
 \=BLAHBL\= AAAAA\=\kill
\> (TI) \> $\bel{\Psi + A}$ \>  $=\bel{\Psi \ast A}$, if $\neg A\notin\bel{\Psi}$\ \\[0.1cm]
\>  \>  \>  $=L$, otherwise\ \\[-0.25em]
\end{tabbing} 
\vspace{-0.75em}

\noindent \LI~ can of course alternatively be presented as $\bel{\Psi \ast A} = \bel{(\Psi \contract \neg A)+A}$. Note that, given \HI~and \LI, the constraint $\mods{\bel{\Psi\ast A}} = \min(\preccurlyeq_\Psi, \mods{A})$ is equivalent to $\mods{\bel{\Psi\contract \neg A}} = \min(\preccurlyeq_\Psi, W) \cup \min(\preccurlyeq_\Psi, \mods{A})$, so that $\preccurlyeq_\Psi$ equally represents both revision and contraction dispositions.

The motivation for \LI~is the following: The most parsimonious way of modifying $\bel{\Psi}$ so as to include $A$ is to simply add the joint logical consequences of $\bel{\Psi}$ and $A$. However, $\Cn(\bel{\Psi}\cup \{A\})$ needn't be consistent. Hence we first `make room' for $A$ by considering instead the belief set $\bel{\Psi\contract \neg A}$ that results  from making the relevant minimal change necessary to achieve consistency.

\subsection{Iterated change}

\begin{centering}
\begin{figure}

\includegraphics{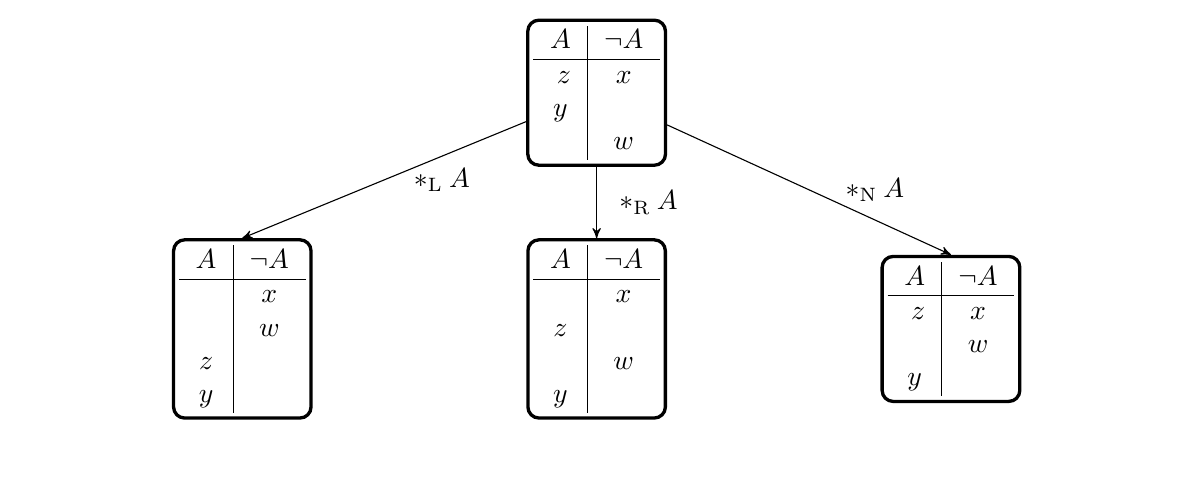}

\vspace{-1em}

\caption{Elementary revision by $A$. The boxes represent states and associated TPOs. The  lower case letters, which represent worlds, are arranged in such a way that the lower the letter, the lower the corresponding world in the relevant ordering. The columns group worlds according to the sentences that they validate. So, for example, in the initial ordering, we have $w \prec y \prec x \sim
z$, with $y, z \in \mods{A}$ and $x, w \in \mods{\neg A}$ and then, after lexicographic revision by $A$, $y \prec z \prec w \prec x$.}
\label{fig:RevProp}
\end{figure}
\end{centering}

In terms of iterated revision, we shall considerably simplify the discussion by restricting our attention to the three principal operators found in the literature. These are natural revision \cite{boutilier1996iterated}:
\begin{itemize}

\item[] $x \preccurlyeq_{\Psi \astN A} y$ iff (1)  $x \in \min(\preccurlyeq_{\Psi}, \mods{A})$, or
(2) $x, y \notin \min(\preccurlyeq_{\Psi}, \mods{A})$ and $x \preccurlyeq_{\Psi} y$

\end{itemize}
restrained revision \cite{booth2006admissible}:
\begin{itemize}

\item[] $x \preccurlyeq_{\Psi \astR A}  y$ iff (1) $x \in \min(\preccurlyeq_{\Psi}, \mods{A})$, or (2)  $x, y \notin \min(\preccurlyeq_{\Psi}, \mods{A})$ and either (a) $x \prec_{\Psi} y$ or (b) $x \sim_{\Psi} y$ and ($x\in\mods{A}$ or $y\in\mods{\neg A}$)

\end{itemize}
and lexicographic revision \cite{nayak2003dynamic}:
\begin{itemize}

\item[] $x \preccurlyeq_{\Psi \astL A}  y$ iff (1)  $x\in\mods{A}$ and $y\in\mods{\neg A}$, or (2) ($x\in\mods{A}$ iff $y\in\mods{A}$) and $x \preccurlyeq_{\Psi} y$.\footnotemark 

\end{itemize}
See Figure \ref{fig:RevProp}.
\footnotetext{These are three of the four iterated revision operators mentioned in Rott's influential survey \cite{Rott2009-ROTSPS}. The remaining operator the irrevocable revision operator of \cite{DBLP:journals/ndjfl/Segerberg98}, which has the unusual characteristic of ensuring that prior  inputs to revision are retained  in the belief set after {\em any} subsequent revision. %For a broader range of operators, see Section 5.2 of  \cite{DBLP:journals/jphil/FermeH11a}.
}

\vspace{-1em}

\noindent All three suggestions operate on the assumption that a state $\Psi$ is to be identified with its corresponding TPO $\preccurlyeq_{\Psi}$ and that belief change functions map pairs of TPOs and sentences onto TPOs,  in other words, they entail: 
\begin{tabbing}
 \=BLAHBL\=\kill
\> \REDRev{} \> If $\preccurlyeq_{\Psi} = \preccurlyeq_{\Psi'}$, then, for any  $A$,  $\preccurlyeq_{\Psi\ast A} = \preccurlyeq_{\Psi'\ast A}$\\[-0.25em]
\end{tabbing} 
\vspace{-0.75em}

\noindent The proposals ensure that $\ast$ satisfies the postulates of Darwiche \& Pearl \cite{darwiche1997logic}. In their semantic forms,  these are: 
\begin{tabbing}
 \=BLAHBL\=\kill
\> \CRevR{1}  \> If $x,y \in \mods{A}$ then $x \preccurlyeq_{\Psi\ast A} y$ iff  $x \preccurlyeq_{\Psi} y$\\ [0.1cm]
\> \CRevR{2} \> If $x,y \in \mods{\neg A}$ then $x \preccurlyeq_{\Psi\ast A} y$ iff  $x \preccurlyeq_{\Psi} y$\\ [0.1cm]
\> \CRevR{3} \> If $x \in \mods{A}$, $y \in \mods{\neg A}$ and $x \prec_{\Psi} y$, then $x \prec_{\Psi\ast A} y$ \\ [0.1cm]
\> \CRevR{4} \> If $x \in \mods{A}$, $y \in \mods{\neg A}$ and $x \preccurlyeq_{\Psi} y$, then $x \preccurlyeq_{\Psi\ast A} y$ \\[-0.25em]
\end{tabbing} 
\vspace{-0.75em}
%\noindent and
%\noindent Syntactically:
%\begin{tabbing}
% \=BLAHBL\=\kill
%\> \CRevS{1} \> If $A \in \mbox{Cn}(B)$ then $\bel{(\Psi \ast A) \ast B} = \bel{\Psi \ast B}$  \\ [0.1cm]
%
%\> \CRevS{2} \> If $\neg A \in \mbox{Cn}(B)$ then $\bel{(\Psi \ast A) \ast B} = \bel{\Psi \ast B}$ \\ [0.1cm]
%
%\> \CRevS{3} \> If $A \in \bel{\Psi \ast B}$ then $A \in \bel{(\Psi \ast A) \ast B}$  \\ 
%
%\> \CRevS{4} \> If $\neg A \not\in \bel{\Psi \ast B}$ then $\neg A \not\in \bel{(\Psi \ast A) \ast B}$\\[-0.25em]
%\end{tabbing} 
%\vspace{-0.75em}

\noindent 
%Regarding $+$, we shall take for granted that, when restricted to inputs consistent with the initial belief set, it also satisfies 
%
Regarding $\contract$, we assume that it satisfies the postulates of Chopra et al \cite{chopra2008iterated}, given  semantically by:
\begin{tabbing}
 \=BLAHBL\=\kill
\> \CConR{1}  \>  If $x,y \in \mods{\neg A}$ then $x \preccurlyeq_{\Psi\contract A} y$ iff  $x \preccurlyeq_{\Psi} y$ \\ [0.1cm]

\> \CConR{2} \> If $x,y \in \mods{A}$ then $x \preccurlyeq_{\Psi\contract A} y$ iff  $x \preccurlyeq_{\Psi} y$\\ [0.1cm]

\> \CConR{3}  \> If $x \in \mods{\neg A}$, $y \in \mods{A}$ and $x \prec_{\Psi} y$ then $x \prec_{\Psi\contract A} y$ \\ [0.1cm]

\> \CConR{4} \> If $x \in \mods{\neg A}$, $y \in \mods{A}$ and $x \preccurlyeq_{\Psi} y$ then $x \preccurlyeq_{\Psi\contract A} y$\\[-0.25em]
\end{tabbing} 
\vspace{-0.75em}
%\begin{tabbing}
% \=BLAHBL\=\kill
%\> \CConS{1}   \>  If $\neg A \in \mbox{Cn}(B)$ then $\bel{(\Psi \contract A) \ast B} = \bel{\Psi \ast B}$ \\ [0.1cm]
%
%\> \CConS{2} \> If $A \in \mbox{Cn}(B)$ then $\bel{(\Psi \contract A) \ast B} = \bel{\Psi \ast B}$ \\ [0.1cm]
%
%\> \CConS{3}  \> If $\neg A \in \bel{\Psi \ast B}$ then $\neg A \in \bel{(\Psi \contract A) \ast B}$ \\ [0.1cm]
%
%\> \CConS{4} \> If $A \not\in \bel{\Psi \ast B}$ then $A \not\in \bel{(\Psi \contract A) \ast B}$\\[-0.25em]
%\end{tabbing} 
%\vspace{-0.75em}

\noindent Concerning  the relations between the belief change operators in the iterated case, we will  be discussing the extension of \LI, as well as that of (TI), later in the paper. Regarding \HI, a proposal for extending the principle to the two-step case was recently floated in \cite{DBLP:conf/ijcai/BoothC16}. Semantically speaking, this involved the characterisation of a binary TPO combination operator $\oplus$, such that $\preccurlyeq_{\Psi\contract A}=\preccurlyeq_{\Psi} \!\oplus\! \preccurlyeq_{\Psi\ast \neg A}$.  Among the baseline constraints on $\oplus$, were a pair of conditions that were shown to be respectively equivalent, in the presence of \CRevR{1} and \CRevR{2}, to the following joint constraints on $\preccurlyeq_{\Psi\contract A}$, $\preccurlyeq_{\Psi}$ and $\preccurlyeq_{\Psi\ast \neg A}$:

\begin{tabbing}
 \=BLAHBL\=\kill
\> \SPU   \> If $x \prec_{\Psi} y$ and $x \prec_{\Psi\ast \neg A} y$ then $x \prec_{\Psi\contract A} y$  \\[0.1cm] 
\> \WPU   \> If $x \preccurlyeq_{\Psi} y$ and $x \preccurlyeq_{\Psi\ast \neg A} y$ then $x \preccurlyeq_{\Psi\contract A} y$  \\[-0.25em]
\end{tabbing} 
\vspace{-0.75em}

\noindent We called operators satisfying such postulates, in addition to \HI, `TeamQueue combinators'. 

\section{Elementary revision operators} 
\label{s:Elementary}

\noindent In this section, we demonstrate the relative generality of the results that follow by providing a characterisation result according to which natural, restrained and lexicographic revision operators are the {\em only} operators satisfying a small set of potentially appealing properties. We shall call operators that satisfy these properties {\em elementary} revision operators. 
 We define elementary revision operators semantically by:
{
\sloppy
\begin{definition}  
$\ast$ is an {\em elementary} revision operator iff it satisfies \CRevR{1}-\CRevR{4}, \IIAPriorRev, \IIAInputRev~and \NeutralityRev.
%\footnote{
%One could also consider elementary {\em contraction} operators, whose characteristic properties are obtained by substituting the postulates of Chopra {\em et al} for the DP postulates. It would appear that one cam show, adapting the proof of Proposition \ref{CharLexRestNat}, that these are limited to: (1) natural contraction, (2) priority, aka `moderate', contraction (see Nayak {\em et al} \hyperlink{Nayaketal05}{2005}, \hyperlink{Nayaketal06}{2006} for both varieties of operator) and (3) a contraction operator that stands to restrained revision as priority contraction stands to lexicographic revision, which, to the best of our knowledge, is new to the literature. However, note that neither (2) nor (3) are consistent with \iHI~(see Section 5.2 of (Booth \& Chandler \hyperlink{BCms}{ms}) for an explanation in connection with (2); the comments carry over to (3)).
%}

\end{definition} 
}

\noindent We have already introduced \CRevR{1}-\CRevR{4}. The remaining principles are new. We  call the first of these `Independence of Irrelevant Alternatives with respect to the prior TPO', after an analogous precept in social choice. For this, we first define the notion of `agreement' between TPO's on a pair of worlds:

\begin{definition}  
Where $\preccurlyeq_{\Psi}, \preccurlyeq_{\Psi'}\in~\TPO{W}$, $\preccurlyeq_{\Psi}$ and $\preccurlyeq_{\Psi'}$ {\em agree} on $\{x, y\}$ iff \\
$\preccurlyeq_{\Psi}\cap\{x,y\}^2 =~\preccurlyeq_{\Psi'}\cap\{x,y\}^2$.

\end{definition} 
\noindent then offer:

\begin{tabbing}
 \=BLAHBL\=\kill
\> \IIAPriorRev \>  If $x, y \notin \min(\preccurlyeq_{\Psi}, \mods{A})\cup \min(\preccurlyeq_{\Psi'}, \mods{A})$, then,  if $\preccurlyeq_{\Psi}$ and $\preccurlyeq_{\Psi'}$ agree on  \\ 
\>\>  $\{x, y\}$, so do $\preccurlyeq_{\Psi\ast A}$ and $\preccurlyeq_{\Psi'\ast A}$\\[-0.25em] 
\end{tabbing} 
\vspace{-0.75em}

%\noindent  We note that this principle is a strengthening of \REDRev{}. Indeed, assume that $\preccurlyeq_{\Psi}=\preccurlyeq_{\Psi'}$. Consider arbitrary $x, y\in W$. If either $x$ or $y$ is in $\min(\preccurlyeq_{\Psi}, \mods{A})\cup \min(\preccurlyeq_{\Psi'}, \mods{A})$, then  $\preccurlyeq_{\Psi\ast A}$ and $\preccurlyeq_{\Psi'\ast A}$ agree on $\{x, y\}$ by virtue of AGM. If $x, y \notin \min(\preccurlyeq_{\Psi}, \mods{A})\cup \min(\preccurlyeq_{\Psi'}, \mods{A})$, then they agree by virtue of
% \IIAPriorRev. Hence $\preccurlyeq_{\Psi\ast A}=\preccurlyeq_{\Psi'\ast A}$, as required.
% \footnote{By virtue of the symmetry of its antecedent, this claim could be equivalently stated with the weaker consequent: $x \preccurlyeq_{\Psi\ast A} y$ iff $x \preccurlyeq_{\Psi'\ast A} y$. Similar comments apply to \IIAInputRev~below.
%}
%$^{,~}$\footnote{A natural question here is whether replacing \IIAPriorRev~by \REDRev{}~in the definition of elementary operators would result in a strictly larger class. \jake{Example of a non-elementary operator that fits the bill}}

\noindent The second new principle--`Independence of Irrelevant Alternatives with respect to the input'--is formally similar to the first. For this we first introduce  some helpful notation:

\begin{definition}
(i) $x\preccurlyeq^A y$ iff $x\in\mods{A}$ or $y\in\mods{\neg A}$, (ii) $x\sim^A y$ when $x\preccurlyeq^A y$ and $y\preccurlyeq^A x$, and (iii) $x\prec^A y$ when $x\preccurlyeq^A y$ but not $y\preccurlyeq^A x$.
\end{definition} 

\noindent The principle is then given by: 

\begin{tabbing}
 \=BLAHBL\=\kill
\> \IIAInputRev\>  If $x, y \notin \min(\preccurlyeq_{\Psi}, \mods{A})\cup \min(\preccurlyeq_{\Psi}, \mods{B})$, then, if  $\preccurlyeq^A$ and $\preccurlyeq^B$ agree on  \\ 
\>\>   $\{x,y\}$, so do $\preccurlyeq_{\Psi\ast A}$ and $\preccurlyeq_{\Psi\ast B}$\\[-0.25em]
\end{tabbing} 
\vspace{-0.75em}

%
%The two above principles can be jointly reformulated as:
%
%\begin{tabbing}
% \=BLAHBL\=\kill
%\> \IIARev \>  If $x, y \notin \bigcup_i \bigcup_j \min(\preccurlyeq_i, \mods{A_j})$, with $i\in\{1,2\}$,\\ 
%\>\> then if (1) $\preccurlyeq^{A_1}$ and $\preccurlyeq^{A_2}$ agree on $x$  \\  
%\>\> and $y$ and (2) so do $\preccurlyeq_1$ and $\preccurlyeq_2$, then (3) so\\  
%\>\> do ${\preccurlyeq_1}^\ast_{A_1}$ and ${\preccurlyeq_2}^\ast_{A_2}$\\[-0.25em]
%\end{tabbing} 
%\vspace{-0.75em}
%
%\noindent \jake{Double-check this is correct!} 

\noindent Although this principle is new to the literature, we note that it can be shown to be equivalent, under our assumptions, to the conjunction of a pair of principles that  were recently defended in \cite{DBLP:conf/kr/0001C18}, where it was shown that they respectively strengthen \CRevR{3} and \CRevR{4}: 

\begin{restatable}{prop}{IIAInputandBetas}
\label{IIAInputandBetas}
Given \CRevR{1}--\CRevR{4}, \IIAInputRev~is equivalent to the conjunction of:
\begin{tabbing}
\=BLAHBLIA \=\kill
\> \BetaRevR{1} \>  If $x \not\in \min(\preccurlyeq_{\Psi}, \mods{C})$, $x\prec^A y$, and    $y \preccurlyeq_{\Psi \ast A} x$,  then $y \preccurlyeq_{\Psi \ast C} x$ \\

\> \BetaRevR{2}  \> If $x \not\in \min(\preccurlyeq_{\Psi}, \mods{C})$, $x\prec^A y$, and $y \prec_{\Psi \ast A} x$, then $y \prec_{\Psi \ast C} x$ \\[-0.25em]
\end{tabbing} 
\vspace{-1.5em} 
\end{restatable}

\noindent The final principle is a principle of `Neutrality', again named after an analogous condition in social choice. To the best of our knowledge, it appears here for the first time in the context of belief revision. Its presentation makes use of the following concept:

\begin{definition}  
Where $A\in L$, $\pi$ is an {\em $A$-preserving order isomorphism} from $\langle W, \preccurlyeq_{\Psi}, \preccurlyeq^{A}\rangle$ to  $\langle W, \preccurlyeq_{\Psi'}, \preccurlyeq^{A}\rangle$  iff  it is a 1:1 mapping from $W$ onto itself such that
\begin{itemize}
\item[(i)] $x \preccurlyeq_{\Psi} y$ iff $\pi(x) \preccurlyeq_{\Psi'} \pi(y)$, and
\item[(ii)] $x \preccurlyeq^{A} y$ iff $\pi(x) \preccurlyeq^{A} \pi(y)$
\end{itemize}

\end{definition} 

\noindent and proceeds as follows:

\begin{tabbing}
 \=BLAHBL\=\kill
\> \NeutralityRev \>  $x \preccurlyeq_{\Psi\ast A} y$ iff $\pi(x) \preccurlyeq_{\Psi'\ast A} \pi(y)$, for any  $A$-preserving order isomorphism  $\pi$ \\
\>\>  from $\langle W, \preccurlyeq_{\Psi}, \preccurlyeq^{A}\rangle$ to $\langle W, \preccurlyeq_{\Psi'}, \preccurlyeq^{A}\rangle$ \\[-0.25em]
\end{tabbing} 
\vspace{-0.75em}

%
%\begin{tabbing}
% \=BLAHBL\=\kill
%\> \NeutralityRev \>  $x \preccurlyeq_{\Psi\ast A} y$ iff $\pi(x) \preccurlyeq'^\ast_A \pi(y)$, for any $(\preccurlyeq^A, \preccurlyeq)$-automorphism  $\pi$ from $\langle W, \preccurlyeq\rangle$ to $\langle W, \preccurlyeq'\rangle$ \\[-0.25em]
%\end{tabbing} 
%\vspace{-0.75em}
%
%\noindent where
%
%\begin{definition}  
%Where $S \subseteq W$ and $A\in L$, $\pi$ is a $(\preccurlyeq_1, \preccurlyeq_2)${\em -automorphism} from $\langle W, \preccurlyeq_1, \preccurlyeq_2\rangle$ to $\langle W, \preccurlyeq_1, \preccurlyeq_2\rangle$ iff  it is a 1:1 mapping from $W$ onto itself such that
%\begin{itemize}
%\item[(i)] $x \preccurlyeq_{\Psi} y$ iff $\pi_A(x) \preccurlyeq_{\Psi'}\pi_A(y)$, and
%\item[(ii)] $x\in S$ iff $\pi_S(x)\in S$
%\end{itemize}
%
%\end{definition} 

%\noindent Taken together, \IIAPriorRev~and \NeutralityRev~are rather strong: they state that, setting aside those pairs of states of which one is minimal with respect to the input to revision $A$, the posterior relation between two worlds depends only on (a) their prior relation and (b) whether or not they respectively validate $A$. \jake{expand here}

\noindent \IIAPriorRev~and \IIAInputRev~say that the relative ordering of $x$ and $y$ after revising by $A$ depends on only (i) their relative order prior to revision (from \IIAPriorRev) and (ii)  their relative positioning with respect to $A$ (i.e., whether or not they satisfy $A$) {\em unless} one of $x$ or $y$ is a minimal $A$-world, in which case this requirement acquiesces to the Success postulate (from  \IIAInputRev).  \NeutralityRev~is a form of language-independence property, stating that the labels (in terms of valuations) of worlds are irrelevant in determining the posterior TPO. The prima facie appeal of these principles is similar to that of their analogues in social choice, substituting a doxastic interpretation of the ordering for a preferential one.

With this in hand, we can now report that:

\begin{restatable}{thm}{CharLexRestNat}
\label{CharLexRestNat}
The only elementary revision operators are lexicographic, restrained and natural revision.
\end{restatable}

\noindent \IIAPriorRev~significantly weakens a principle introduced under the name of `(IIA)' in \cite{Glaister1998}, which simply corresponds to the embedded conditional: If $\preccurlyeq_{\Psi}$ and $\preccurlyeq_{\Psi'}$ agree on $\{x, y\}$, so do $\preccurlyeq_{\Psi\ast A}$ and $\preccurlyeq_{\Psi'\ast A}$.  \IIAInputRev~amounts to a similar weakening of a condition found in \cite{booth2011revise}. An interesting question, therefore, arises as to why the stronger principles do not figure in our characterisation.

The unqualified version of \IIAPriorRev~is  only satisfied by $\astL$, assuming \CRevR{1} and \CRevR{2}~and that the domain of the revision function is $\TPO{W}$. Indeed, let $x\in\mods{A}$ and $y\in\mods{\neg A}$. Then, for any $\preccurlyeq_{\Psi}$, there will exist $\preccurlyeq_{\Psi'}$ in $\TPO{W}$ that agrees with $\preccurlyeq_{\Psi}$ on $\{x, y\}$ and is such that $x\in \min(\preccurlyeq_{\Psi'}, \mods{A})$ (and, since $y\in\mods{\neg A}$, $y\notin \min(\preccurlyeq_{\Psi'}, \mods{A})$). But by AGM, if $x\in \min(\preccurlyeq_{\Psi'}, \mods{A})$ but $y\notin \min(\preccurlyeq_{\Psi'}, \mods{A})$, then $x\prec_{\Psi' \ast A} y$. So, by the unqualified version of \IIAPriorRev, $x\prec_{\Psi \ast A} y$. Hence, if $x\in\mods{A}$ and $y\in\mods{\neg A}$, then $x\prec_{\Psi \ast A} y$, a condition  only satisfied by $\astL$, assuming \CRevR{1} and \CRevR{2}.\footnote{We note that \cite{Glaister1998} offers a rather different characterisation of lexicographic revision that also involves the unqualified version of \IIAPriorRev. 
%See his \jake{theorem ??}. However, this result also makes use of \jake{something super weird; explain}.
}

Similarly,  Booth \& Meyer's strong version of \IIAInputRev, in conjunction with \CRevR{1}--\CRevR{4}, can be shown to entail a principle that we have called  `\BetaRevPlusR{1}' in previous work \cite{DBLP:conf/kr/0001C18}, where we showed (see Corollary 1 there) to characterise lexicographic revision, given AGM and \CRevR{1}-\CRevR{2}.\footnote{\label{IIAOriginal} In the proof of Proposition \ref{IIAInputandBetas}  above, we established the equivalence between \IIAInputRev~and the conjunction of \BetaRevR{1} and \BetaRevR{2} using only \CRevR{1}-- \CRevR{4}. This proof can be adapted to establish a strengthening of Booth \& Meyer's Proposition 3 in which, unlike in the original, the principle of `Independence'  \IndRevR~is not appealed to. In other words: In the presence of AGM and \CRevR{1}-- \CRevR{4}, Booth \& Meyer's strengthening of \IIAInputRev~ is equivalent to the conjunction of what \cite{DBLP:conf/kr/0001C18} call `\BetaRevPlusR{1}' and `\BetaRevPlusR{2}'.}

\section{Extending the Levi Identity}
\label{s:iLI}

\subsection{A proposal involving rational closure}
\label{ss:ThreeRed}

The most straightforward syntactic extension of \LI~would involve replacing all belief sets by conditional belief sets, leaving all else unchanged. This would require extending the domain of $\Cn$ to subsets of the conditional  language $L_c$, which can be naturally achieved by setting, for $\Delta\subseteq L_c$, $\Cn(\Delta)=\Delta\cup \Cn(\Delta\cap L)$. So we would be considering the claim that $\cbel{\Psi \ast A} = \mathrm{Cn}(\cbel{\Psi \contract \neg A} \cup \{A\})$. 
%:
%\begin{tabbing}
% \=BLAHBL\=\kill
%\> (iLI2) \> $\cbel{\Psi \ast A} = \mathrm{Cn}(\cbel{\Psi \contract \neg A} \cup \{A\})$ \\[-0.25em]
%\end{tabbing} 
%\vspace{-0.75em}
%\noindent 
%
This, however, is a bad idea, since it is easy to show that:

\begin{restatable}{prop}{iLITwoTriv}
\label{iLITwoTriv}
If $\cbel{\Psi \ast A} = \mathrm{Cn}(\cbel{\Psi \contract \neg A} \cup \{A\})$, then there are no consistent belief sets, given the two following AGM postulates:
\begin{tabbing}
\=BLAHBLI: \=\kill

\> \KRev{2} \> $A\in [\Psi\ast  A]$\\[0.1cm]

\> \KCon{2}  \> $[\Psi\contract  A]\subseteq[\Psi]$\\[-0.25em]

\end{tabbing}
\vspace{-1.5em}
\end{restatable}

%\footnotetext{There is of course another obvious naive reductive way to extend \LI~syntactically, namely by replacing all instances of belief states with instances of those states revised by some sentence $B$, as in: $\bel{(\Psi \ast A)\ast B} = \mathrm{Cn}(\bel{(\Psi \contract \neg A) \ast  B} \cup \{A\})$. Given AGM, we would recover \LI~as the special case in which $B\in \Cn(\varnothing)$.  But the suggestion is also clearly a non-starter, since it entails that $A \in \bel{(\Psi \ast A) \ast B}$ for all $A, B$ and $\Psi$.}

%\vspace{-1em}

%
%\begin{pproof}
%Assume $\cbel{\Psi \ast A} = \mathrm{Cn}(\cbel{\Psi \contract \neg A} \cup \{A\})$. By Success,  $A\in\bel{\Psi\ast A}$. By the Ramsey Test, $\top\Rightarrow A\in \cbel{\Psi\ast A}$ and so $\top\Rightarrow A\in \mathrm{Cn}(\cbel{\Psi \contract \neg A} \cup \{A\})$. But then, since $\top\Rightarrow A\notin L$ and, as we have stipulated, for $\Delta\subseteq L_c$, $\Cn(\Delta)=\Delta\cup \Cn(\Delta\cap L)$, it must be the case that $\top\Rightarrow A\in \cbel{\Psi \contract \neg A}$. Hence, by the Ramsey Test again, it follows that $A \in \bel{\Psi\contract \neg A}$. From \KCon{2}, we then have $A\in \bel{\Psi}$. By a similar chain of reasoning, we can establish that $\neg A\in \bel{\Psi}$. \qed
%\end{pproof}

%\vspace{1em}

\noindent The core issue highlighted by this result is that the right hand side of the equality won't generally correspond to a rational consequence relation, due to the fact that $\mathrm{Cn}$ simply yields too small a set of consequences.  So a natural suggestion here would be to make use of the rational closure operator $\CRat$ of \cite{lehmann1992does} instead of $\mathrm{Cn}$. Indeed, $\CRat$ has been touted as offering the appropriately  conservative way of extending a set of conditionals to something that corresponds to a rational consequence relation (see \cite{lehmann1992does}).
%, but also \cite{Rott1997-ROTDIF}). %and \cite{DBLP:journals/logcom/HillP03}. 
This  gives us the `iterated Levi Identity using Rational Closure' (or `\iLIR' for short):

\begin{tabbing}
 \=BLAHBL\=\kill
\> \iLIR \> $\cbel{\Psi \ast A} = \CRat(\cbel{\Psi \contract \neg A} \cup \{A\})$\footnotemark \\[-0.25em]
\end{tabbing} 
\vspace{-1.5em}

\footnotetext{Strictly speaking, $\CRat$~is an operation on {\em purely} conditional belief sets. However, it can be obviously generalised to the case in which the set includes non-conditionals, since for any $A\in L$, $A\in\cbel{\Psi}$ iff $\top\Rightarrow A\in\cbel{\Psi}$.}

%\noindent  We also recover a conditional belief set that {\em does} correspond to a rational consequence relation. So we are in more promising territory than we were in relation to the two initial suggestions.

%\jake{Cut this paragraph + Prop 4 out in favour of a more general result that only uses TQ?} We note that \iLIR~is reminiscent of the principle \iHI~introduced in Section \ref{s:Preliminaries}, which also appeals to $\CRat$. It isn't immediately obvious that \iLIR~enjoys the same kind of appeal as \iHI. This issue needs to be investigated in greater depth. In any case, it is worth thinking about the result of endorsing \iLIR~alongside \iHI: Would this constitute a sensible combination? Well, the following holds true:
%
%\begin{restatable}{prop}{iLIRandiHIRNatLexRest}
%\label{iLIRandiHIRNatLexRest}
%Natural revision is the only elementary revision operator $\ast$ for which there exists a contraction operator $\contract$ such that $\ast$ and $\contract$ satisfy both  \iLIR~and \iHI.
%\end{restatable}

\subsection{Nayak {\em et al}'s `New Levi Identity'}
\label{ss:NLI}

An alternative extension of \LI~can be obtained by using an iterable expansion operator +. This is  the `New Levi Identity'  of Nayak {\em et al}, which  is briefly presented in  \cite{DBLP:conf/dagstuhl/NayakGOP05}. Semantically, it is given by:

 \begin{tabbing}
 \=BLAHBL\=\kill
\> \NLI \> $\preccurlyeq_{\Psi\ast A} = \preccurlyeq_{(\Psi\contract \neg A) + A}$\\[-0.25em]
\end{tabbing} 
\vspace{-0.75em}

\noindent  Syntactically, in terms of conditional belief sets, we then would have: $\cbel{\Psi\ast A} = \cbel{(\Psi\contract \neg A)+ A}$. 

It is easily verified  that \LI~follows from \NLI, if one assumes, for instance, that $\contract$ satisfies \CConR{1}. Indeed, \LI~amounts to $\min(\preccurlyeq_{\Psi}, \mods{A}) = \min(\preccurlyeq_{\Psi \contract \neg A}, W)\cap \mods{A}= \min(\preccurlyeq_{\Psi \contract \neg A}, \mods{A})$, which immediately follows from \CConR{1}. \NLI~also has some other interesting general properties. For example, one can show, rather trivially, that:

\begin{restatable}{prop}{ChopratoDP}
\label{ChopratoDP}
If $\ast$ and $\contract$ satisfy \NLI, then, for $i\in\{1,2,3,4\}$,  \CConR{i} entails \CRevR{i}, if $+$ also satisfies \CRevR{i}. 
%\jake{for $\preccurlyeq_{\Psi}$ such that $\min(\preccurlyeq_{\Psi}, W)\nsubseteq \neg A$?}
\end{restatable}

\noindent  This result mirrors a result in \cite{DBLP:conf/ijcai/BoothC16}, in which it was shown that TeamQueue  combination  allows one to move from each \CRevR{i} to the corresponding \CConR{i}.

Assuming,  as  Nayak {\em et al} do, the following natural semantic iterated version of (TI): 

 \begin{tabbing}
 \=BLAHBL\= AAIAI\=\kill
\> (iTI$_{\preccurlyeq}$) \> $\preccurlyeq_{\Psi + A}$ \>  $=\preccurlyeq_{\Psi\ast A}$, if $\min(\preccurlyeq, W)\nsubseteq\mods{\neg A}$\ \\[0.1cm]
\>  \>  \>  $=\preccurlyeq_{\Psi_\bot}$, otherwise\ \\[-0.25em]
\end{tabbing} 
\vspace{-0.75em}

\noindent where $\Psi_\bot$ is an `absurd' epistemic state such that $\bel{\Psi_\bot} = L$,\footnote{Nayak {\em et al} have little to say about $\Psi_\bot$, aside from its being the case that $\preccurlyeq_{\Psi_\bot \contract A}$  is such that $x\sim_{\Psi_\bot \contract A} y$ for all $x, y\in W$. More recently,  \cite{Ferme2018} have suggested that the state resulting from expansion into inconsistency be defined in a more fine-grained manner, in a proposal that involves introducing an `impossible' world such that $w_\bot\models A$ for all $A\in L$. We refer the reader to their paper for further details, since nothing here hinges on the distinction between these views.} \NLI~is equivalent to:
%\footnote{Although it then turns out that 
% they further assume that $\ast$/$+$ is lexicographic revision, so the discussion isn't even that general. \jake{sort this out}}
 
 \begin{tabbing}
 \=BLAHBL\=\kill
\> \iLINRSem \> $\preccurlyeq_{\Psi\ast A} = \preccurlyeq_{(\Psi\contract \neg A)\ast A}$\\[-0.25em]
\end{tabbing} 
\vspace{-0.75em}

\noindent 
% Equivalently, in terms of unconditional belief sets: $\bel{(\Psi\ast A)\ast B} = \bel{((\Psi\contract \neg A)\ast A)\ast B}$. 
%Importantly, unlike what we have in the previous suggestions, $\ast$ appears on both the left- and right- hand sides of the equality in \NLI. So the principle doesn't offer a {\em reduction} of $\ast$ to $\contract$. 
\noindent In what follows, then, we shall use \NLI~and \iLINRSem~interchangeably. Importantly, while the proposal considered in the previous section was reductive, in the sense that the operator $\ast$ on the left-hand side of the identity did not appear on the right, \iLINRSem~features $\ast$ on both sides.

To date, however, the implications of this principle have not been studied in any kind of detail.  In what follows, we offer some new results of interest. We first  note: 

\begin{restatable}{thm}{AltNecSuffNLIcopy}
\label{AltNecSuffNLIcopy}
If $\ast$ is an elementary revision operator and  $\contract$ satisfies \CConR{1}-\CConR{4}, then $\ast$ and $\contract$ satisfy \NLI~iff they satisfy the following: 
\begin{tabbing}
 \=BLAHBL\=\kill
\> \CConRevR{1} \>   If $x, y\in\mods{A}$, then $x\preccurlyeq_{\Psi \contract \neg A} y$ iff $x\preccurlyeq_{\Psi\ast A} y$ \\[0.1cm] 
\> \CConRevR{2} \>    If $x, y\in\mods{\neg A}$, then $x\preccurlyeq_{\Psi \contract \neg A} y$ iff $x\preccurlyeq_{\Psi\ast A} y$ \\[0.1cm] 
\> \CConRevR{3} \>     If $x\in\mods{A}$, $y\in\mods{\neg A}$ and  $x\prec_{\Psi \contract \neg A} y$, then $x\prec_{\Psi\ast A} y$.  \\ [0.1cm]
\> \CConRevR{4} \>     If $x\in\mods{A}$, $y\in\mods{\neg A}$ and   $x\preccurlyeq_{\Psi \contract \neg A} y$, then $x\preccurlyeq_{\Psi\ast A} y$.   \\[-0.25em]
\end{tabbing}  
\vspace{-1.5em}
\end{restatable}

%\begin{proofsketch}
%Each direction is proven separately. Regarding the right-to-left direction:
%
%\begin{restatable}{lem}{SuffNLI}
%\label{SuffNLI}
%If $\ast$ is an elementary revision operator, $\contract$ satisfies \CConR{1}-\CConR{4},  and $\contract$ and $\ast$ satisfy \CConRevR{1}-\CConRevR{4}, then $\ast$ and $\contract$ satisfy \NLI. 
%\end{restatable}
%
%\noindent Regarding the  left-to-right direction, we actually prove the following stronger claim:
%
%\begin{restatable}{lem}{ConRevNec}
%\label{ConRevNec}
%If $\ast$ satisfies \CRevR{1}--\CRevR{4}, then there exists $\contract$ such that $\ast$ and $\contract$ satisfy \NLI~only if $\ast$ and $\contract$ satisfy \CConRevR{1}--\CConRevR{4}.\qed
%\end{restatable}
%
%
%\end{proofsketch}

\noindent The principles \CConRevR{1}-\CConRevR{4} are new to the literature and bear an obvious formal resemblance to the postulates of Darwiche \& Pearl and of Chopra {\em et al}. Taken together, they require contraction by $\neg A$ to be a kind of `{\em mild revision}' by $A$, since they tell us that the position of any $A$-world with respect to any $\neg A$-world is at least as good after revision by $A$ as it is after contraction by $\neg A$.

Somewhat surprisingly (to us), it turns out that these principles are {\em also} closely connected to  the semantic `TeamQueue combinator'  approach to extending the Harper Identity to the iterated case that was proposed in \cite{DBLP:conf/ijcai/BoothC16}. 
Indeed, one can show that:

\begin{restatable}{thm}{SWPUandConRev}
\label{SWPUandConRev}
If $\ast$ satisfies \CRevR{1}-\CRevR{4} and $\contract$ satisfies \CConR{1}-\CConR{4}, then $\ast$ and $\contract$ satisfy \CConRevR{1}-\CConRevR{4} iff they satisfy \SPU~and \WPU.
\end{restatable}

\noindent In conjunction with Theorem \ref{AltNecSuffNLIcopy}, Theorem \ref{SWPUandConRev} entails:

\begin{restatable}{cor}{NLIandTQ}
\label{NLIandTQ}
If $\ast$ is an elementary revision operator and  $\contract$ satisfies \CConR{1}-\CConR{4}, then $\ast$ and $\contract$ satisfy \NLI~iff they satisfy \SPU~and \WPU.
\end{restatable}

\noindent In this particular context, then, \NLI~simply amounts to the conjunction of a pair of constraints proposed in the context of extending \HI~to the iterated case.

\subsection{Rational closure and the New Levi Identity}
\label{ss:RCNLI}

At this stage, we have considered both a potentially promising reductive proposal and a promising non-reductive one. A natural question, then, is: How would these two suggestions fare in conjunction with one another? To answer this question, we provide the semantic counterpart for  our first principle, which was formulated only syntactically:

\begin{restatable}{thm}{LeviIDRatNat}
\label{LeviIDRatNat}
Given AGM, \iLIR~is equivalent to :
\begin{tabbing}
 \=BLAHBLAA\=\kill
\> \iLIRSem \> $\preccurlyeq_{\Psi \ast A} = \preccurlyeq_{(\Psi \contract \neg A) \astN A}$\footnotemark{}\\[-0.25em]
\end{tabbing} 
\vspace{-1.5em}
\end{restatable}

\footnotetext{Note that \cite{10.1007/978-3-319-67582-4_25} explicitly mention  \iLIRSem~and flag it out as a potentially desirable principle. 
%The reason for their tentative approval appears to be the following: For any contraction operator satisfying the postulates of Chopra et al, \iLIRSem will yield a revision operator that satisfies the postulates of Darwiche \& Pearl. \jake{???}
}

{\sloppy
\noindent With this in hand, the consequences of endorsing \iLIR\, on the heels of \NLI\, should be obvious:  rational iterated revision would have to coincide with natural revision.
% \jake{Add comment, suggested by 3rd reviewer, that this doesn't bode well for \iLIR.}
}

This raises an interesting question: For each remaining elementary operator $\ast$, does there exist a suitable alternative closure operator $\C$, such that  $\cbel{\Psi \ast A} = \C(\cbel{\Psi \contract \neg A} \cup \{A\})$ iff  $\preccurlyeq_{\Psi \ast A} = \preccurlyeq_{(\Psi \contract \neg A) \ast A}$?\footnote{Note the importance of \REDRev{} in making this kind of correspondence  even {\em prima facie} possible. Indeed, if \REDRev{} fails, then $\preccurlyeq_{\Psi \contract \neg A}$ and $A$ will fail to jointly determine $\preccurlyeq_{\Psi \ast A}$. In syntactic terms, $\cbel{\Psi \contract \neg A}$ and $A$ will fail to jointly determine $\cbel{\Psi \ast A}$. }  Indeed, although rational closure is by far the most popular closure operator in the literature, alternative closure operators have been proposed,  including, for instance  the lexicographic closure operator of \cite{Lehmann92anotherperspective} or again the maximum entropy closure operator of \cite{DBLP:conf/aaai/GoldszmidtMP90}. Furthermore, there has been some limited work on potential  connections between closure operators and revision operators (namely \cite{Booth2001}). However, this work has only focussed on the relation between lexicographic closure and lexicographic revision and its pertinence to the current problem remains unclear. 

Although we do not currently have an answer to our question, we can report that the existence of suitable relevant closure operators will very much depend on the manner in which one extends \HI~to the iterated case. To illustrate, in a previous discussion of  the issue \cite{DBLP:conf/ijcai/BoothC16}, we considered a particular TeamQueue combinator, $\STQ$.  We showed, in Section 6 of that paper, that for $\ast=\astL$ or $\ast=\astR$, the equality $\preccurlyeq_{\Psi\contract A}=\preccurlyeq_{\Psi} \!\STQ\! \preccurlyeq_{\Psi\ast \neg A}$ entails that $\contract = \contract_{\mathrm{STQL}}$, where $\contract_{\mathrm{STQL}}$ is an iterated contraction operator that we call `STQ-Lex'. We can, however, show the following: 

\begin{restatable}{prop}{ClosureImp}
\label{ClosureImp}
If  $\ast=\astL$ or $\ast=\astR$ and $\contract = \contract_{\mathrm{STQL}}$, then there exists no closure operator $\C$, satisfying the property of Rational Identity: 
 \begin{tabbing}
 \=BLAHBL \=\kill
%\> (Inc) \> $\Delta\subseteq\C(\Delta)$.\\[0.1cm]
%\> (RID) \> If $\Delta$ is rational, then $\C(\Delta) \subseteq \Delta$. \\[-0.25em]
\> \RID \> If $\Delta$ is rational, then $\C(\Delta) = \Delta$. \\[-0.25em]
\end{tabbing} 
\vspace{-0.75em}
such that both \NLI~and $\cbel{\Psi \ast A} = \mathrm{C}(\cbel{\Psi \contract \neg A} \cup \{A\})$ are true. 
\end{restatable}

\noindent \RID~seems a desirable property of closure operators, which aim to extend a set of conditionals $\Delta$ to that rational set of conditionals whose endorsement is mandated by that of  $\Delta$. The standard postulate of  Inclusion ($\Delta\subseteq \C (\Delta)$) tell us that $\C$ must extend $\Delta$ to a rational superset of $\Delta$. \RID~adds to this the notion that if $\Delta$ `ain't broke', it needn't be `fixed'. 

Interestingly, the proof of this impossibility result fails to go through when $\ast=\astL$ and $\contract=\contract_{\mathrm{P}}$, where $\contract_{\mathrm{P}}$ is the priority contraction operator of \cite{DBLP:conf/dagstuhl/NayakGOP05}. In \cite{DBLP:conf/ijcai/BoothC16} we note that priority contraction can be recovered from lexicographic revision via a particular TeamQueue combinator. Furthermore, the same combinator can be used to define a contraction operator from restrained revision (call it $\contract_{\mathrm{R}}$). Again, the proof of the above result breaks down when $\ast=\astR$ and $\contract=\contract_{\mathrm{R}}$.

%=====================================================

\section{Is iterated revision reducible to iterated contraction?}\label{s:reducibility}

%=====================================================

%\jake{One key question, which Nayak {\em et al} do not consider, is the following: Is there a principled reason to restrict one's attention to \NLI~rather than another member of the family? It turns out that we can show that \NLI~is {\em the only member of this family} that is consistent with a certain set of assumptions regarding the operators $\ast$, $+$ and $\contract$. More specifically:
%
%\begin{restatable}{prop}{WhyNLIcopy}
%\label{}
%If $\ast$ and $+$ are elementary \jake{revision} operators, $\contract$ satisfies \CConR{1}-\CConR{4},  and $\contract$ and $\ast$ satisfy \CConRevR{1}-\CConRevR{4}, then $\preccurlyeq_{\Psi\ast A} = \preccurlyeq_{(\Psi \contract \neg A) + A}$ iff $+ = \ast$.
%
%\end{restatable}
%}

Konieczny and Pino P\'erez \cite[Theorem 5]{10.1007/978-3-319-67582-4_25} plausibly claim that, for a finitely generated language, the cardinality of (i) the set of revision operators that satisfy both the AGM postulates for revision and \CRevR{1}--\CRevR{4} is strictly greater than the cardinality of (ii) the set of contraction operators that satisfy both the AGM postulates for contraction and \CConR{1}--\CConR{4}. From this, they conclude that there is no bijection between rational iterated revision and contraction operators and hence no reduction of iterated revision to iterated contraction. 

But this conclusion is not warranted without a further argument to the effect that every member of (i) is rational. In other words, it could be the case that \CRevR{1}--\CRevR{4}  need supplementing. This has certainly been the belief of the proponents of the various elementary revision operators that we have discussed in the present paper. And indeed, the proponent of $\astN$ could claim, endorsing our $\STQ$-based extension of \HI, that rational contraction goes by natural contraction. By the same principle, proponents of $\astR$ or $\astL$ could respectively claim that rational contraction goes by natural contraction or STQ-Lex contraction, respectively (see \cite[Section 6]{DBLP:conf/ijcai/BoothC16}). Those are three candidate bijections that are all consistent, furthermore, with \NLI. 

One could nevertheless run an arguably  plausible  argument to Konieczny and Pino P\'erez's desired conclusion based on the observation that natural and restrained revision are both mapped onto natural contraction by the $\STQ$ method. Even if one thinks that it is implausible to claim that iterated change {\em must} comply with one of either restrained or natural revision, it is not implausible to claim that it sometimes {\em may} comply with either. In other words: There plausibly exists at least one prior TPO that is rationally consistent with two distinct potential posterior TPOs, respectively obtained via natural and restrained revision by a given sentence $A$. Given the $\STQ$-based extension of \HI, only one posterior TPO can be obtained by contraction by $\neg A$, namely the one obtained by natural contraction by $\neg A$. But if this is true, iterated revision dispositions cannot be recovered from iterated contraction dispositions.

%=====================================================

\section{Conclusions and further work}\label{s:Conclusions}

%=====================================================

We have considered two possible extensions of \LI~to the iterated case: a reductive proposal \iLIR~based on the rational closure operator, and a non-reductive proposal \NLI~that involves a contraction step, followed by an expansion. We have shown that, when restricted to a popular class of `elementary' revision operators, \NLI~is in fact equivalent, under weak assumptions,  to both (i) a new set of postulates \CConRevR{1}-\CConRevR{4} and (ii) a pair of principles recently defended in the literature on \HI. 

However,  it has also been noted that   \iLIR~has strong consequences when conjoined with \NLI. This suggests the need for (1) a  future consideration of various alternatives to the former that make use of surrogate closure operators. 

Furthermore, the revision operators of the class that we have focussed on have been criticised for their equation of belief states with TPOs (the principle \REDRev{}; see \cite{DBLP:journals/jphil/BoothC17}). 
%
%In response, \cite{DBLP:conf/kr/0001C18} introduced a more general semantic framework, induced by comparatively  liberal set of rationality constraints. The family of POI revision operators can be represented by operations on a structure called a `proper ordinal interval' assignment. It includes both restrained and lexicographic revision as special cases. As Booth \& Chandler note, while natural revision operators do not fall into this category, they do belong  to a slightly larger  class, associated with operations on a yet more general structure. 
%
One obvious extension of our work would be (2) an exploration of the extent to which the results reported in Section \ref{s:iLI} carry over to operators that avoid this identification, such as the POI operators of \cite{DBLP:conf/kr/0001C18}.

%
%Finally, our characterisation of the elementary operators is of interest in itself. Due to space considerations, however, we have carried the exercise out in entirely semantic terms. A fuller treatment would require (3) a discussion of the syntactic counterparts of the semantic properties presented here.

%
%\vfill
%
%\pagebreak

\bibliographystyle{splncs04}
\bibliography{EIROLILongLORI}

%\vfill
%=====================================================

\section*{Appendix}\label{s:appendix}

%=====================================================

\IIAInputandBetas*

\begin{pproof}
The proof of this claim closely resembles the proof of Proposition 3 of \cite{booth2011revise}. We first establish the following lemma:

\begin{restatable}{lem}{BMstronger}
\label{BMstronger}
Given \CRevR{1}--\CRevR{4},
\begin{itemize}
\item[(a)] If $x\prec^A y$ and $\preccurlyeq^A$ and $\preccurlyeq^C$ do not agree on $\{x, y\}$, then, if $y \preccurlyeq_{\Psi\ast A} x$, then $y \preccurlyeq_{\Psi \ast C} x$. 

\item[(b)] If $x\prec^A y$ and $\preccurlyeq^A$ and $\preccurlyeq^C$ do not agree on $\{x, y\}$, then, if $y \prec_{\Psi\ast A} x$, then $y \prec_{\Psi \ast C} x$. 
\end{itemize}
\end{restatable}

\noindent We simply derive (a), since the proof of (b) is analogous. Assume that $x\prec^A y$ and $\preccurlyeq^A$ and that $\preccurlyeq^C$ do not agree on $\{x, y\}$. In other words: $x\in\mods{A}, y\in\mods{\neg A}$, and either (i) $x\in\mods{C}, y\in\mods{C}$, (ii) $x\in\mods{\neg C}, y\in\mods{\neg C}$ or (iii) $x\in\mods{\neg C}, y\in\mods{C}$. Assume that $y \preccurlyeq_{\Psi\ast A} x$. From this and $x\in\mods{A}, y\in\mods{\neg A}$, it follows, by \CRevR{3}, that $y \preccurlyeq_{\Psi} x$. From this, if either (i), (ii) or (iii) hold, then, by \CRevR{1}, \CRevR{2}, and \CRevR{4}, respectively, we have $y \preccurlyeq_{\Psi \ast C} x$, as required. This completes the proof of Lemma \ref{BMstronger}.

With this in hand, we can derive each direction of the equivalence:
\begin{itemize}

\item[{\bf (a)}] { From \IIAInputRev~to \BetaRevR{1} and \BetaRevR{2}:} Regarding \BetaRevR{1}, assume $x \not\in \min(\preccurlyeq_{\Psi}, \mods{C})$, $x\prec^A y$, and    $y \preccurlyeq_{\Psi \ast A} x$. If $\preccurlyeq^A$ and $\preccurlyeq^C$ do not agree on $\{x, y\}$, then the required result follows by principle (a) of Lemma \ref{BMstronger}. So assume that they do agree, and hence that $x\prec^C y$. We now establish that $x, y \notin \min(\preccurlyeq_{\Psi}, \mods{A})\cup \min(\preccurlyeq_{\Psi}, \mods{C})$. We already have $x \not\in \min(\preccurlyeq_{\Psi}, \mods{C})$. Since, by $x\prec^C y$, it follows that $y\in\mods{\neg C}$, we have  $y \not\in \min(\preccurlyeq_{\Psi}, \mods{C})$. Furthermore, by $x\prec^A y$, it follows that $y\in\mods{\neg A}$ and so $y \not\in \min(\preccurlyeq_{\Psi}, \mods{A})$. Finally, assume for contradiction that $x\in \min(\preccurlyeq_{\Psi}, \mods{A})$. Then $x\in \min(\preccurlyeq_{\Psi \ast A}, W)$. Since $y\in\mods{\neg A}$, by Success, $y\notin \min(\preccurlyeq_{\Psi \ast A}, W)$. Hence $x \prec_{\Psi \ast A} y$, contradicting $y \preccurlyeq_{\Psi \ast A} x$. So we can infer that $x\notin \min(\preccurlyeq_{\Psi}, \mods{A})$. With this in hand, we can apply \IIAInputRev~to derive $y \preccurlyeq_{\Psi \ast C} x$, as required. The derivation of \BetaRevR{2} is analogous.

\item[{\bf (b)}] {From \BetaRevR{1} and \BetaRevR{2} to \IIAInputRev:} Assume that $x, y \notin \min(\preccurlyeq_{\Psi}, \mods{A})\cup \min(\preccurlyeq_{\Psi}, \mods{C})$ and that $\preccurlyeq^A$ and $\preccurlyeq^C$ agree on $\{x, y\}$. We want to show that $x\preccurlyeq_{\Psi \ast A} y$ iff $x\preccurlyeq_{\Psi \ast C} y$. By symmetry, it suffices for this to show that $x\preccurlyeq_{\Psi \ast A} y$ implies $x\preccurlyeq_{\Psi \ast C} y$. So assume $x\preccurlyeq_{\Psi \ast A} y$. Since $\preccurlyeq^A$ and $\preccurlyeq^C$ agree on $\{x, y\}$, we have three cases to consider: 
\begin{itemize}
\item[{\bf (i)}] {$x\prec^A y$ and $x\prec^C y$:} Assume for contradiction that $y\prec_{\Psi \ast C} x$. From this, $x \not\in \min(\preccurlyeq_{\Psi}, \mods{A})$  and $x\prec^C y$, it follows by \BetaRevR{2} that $y\prec_{\Psi \ast A} x$, contradicting $x\preccurlyeq_{\Psi \ast A} y$. Hence $x\preccurlyeq_{\Psi \ast C} y$, as required.

\item[{\bf(ii)}] {$x\sim^A y$ and $x\sim^C y$:} It follows from this, via \CRevR{1} and \CRevR{2}, that $x\preccurlyeq_{\Psi \ast A} y$ iff $x\preccurlyeq_{\Psi} y$ iff $x\preccurlyeq_{\Psi \ast C} y$. Hence $x\preccurlyeq_{\Psi \ast C} y$, as required.

\item[{\bf(iii)}] {$y\prec^A x$ and $y\prec^C x$:} By \BetaRevR{1}, it follows, from $x \not\in \min(\preccurlyeq_{\Psi}, \mods{A})$, $y\prec^A x$, and $x\preccurlyeq_{\Psi \ast A} y$, that $x\preccurlyeq_{\Psi \ast C} y$, as required. \qed
\end{itemize}
\end{itemize}
\end{pproof}

\vspace{1em}

%=======================================================

%
%

%=====================================================

\CharLexRestNat*

\begin{pproof} We prove the result in its two obvious parts. First:

\begin{restatable}{lem}{NRLareElem}
\label{NRLareElem}
Lexicographic, restrained and natural revision operators are elementary operators
\end{restatable}

\noindent It is obvious that \NeutralityRev~is satisfied by satisfied by lexicographic, restrained and natural revision operators. It is also well known that these operators satisfy \CRevR{1}-\CRevR{4}. We quickly verify here that they also satisfy \IIAPriorRev:
\begin{itemize}

\item[{\bf(a)}] {Regarding lexicographic revision:} The principle actually holds without the requirement that $x, y \notin \min(\preccurlyeq_{\Psi}, \mods{A})\cup \min(\preccurlyeq_{\Psi'}, \mods{A})$. We consider 3 cases:
\begin{itemize}

\item[{\bf (i)}] {$x \in \mods{A}$ and $y \in \mods{\neg A}$:} Then $x \prec_{\Psi \ast A} y$ and  $x \prec_{\Psi' \ast A} y$

\item[{\bf (ii)}] {$y \in \mods{A}$ and $x \in \mods{\neg A}$: }Then $y \prec_{\Psi \ast A} x$ and  $y \prec_{\Psi' \ast A} x$

\item[{\bf (iii)}] {$x,y \in \mods{A}$ or $x,y \in \mods{\neg A}$: }Then $x \preccurlyeq_{\Psi} y$ iff $x \preccurlyeq_{\Psi \ast A} y$ and $x \preccurlyeq_{\Psi'}y$ iff $x \preccurlyeq_{\Psi' \ast A} y$. Also: $y \preccurlyeq_{\Psi} x$ iff $y \preccurlyeq_{\Psi \ast A} x$ and $y \preccurlyeq_{\Psi'}x$ iff $y \preccurlyeq_{\Psi' \ast A} x$.

\end{itemize}
\item[{\bf(b)}] {Regarding restrained revision:} Here we consider again 3 cases, this time depending on the prior relation between $x$ and $y$:
\begin{itemize}

\item[{\bf (i)}] {$x \sim_{\Psi} y$ and $x \sim_{\Psi'} y$: }If $x \in \mods{A}$ and $y \in \mods{\neg A}$, then $x \prec_{\Psi \ast A} y$ and  $x \prec_{\Psi' \ast A} y$. Similarly, if $y \in \mods{A}$ and $x \in \mods{\neg A}$, then $y \prec_{\Psi \ast A} x$ and  $y \prec_{\Psi' \ast A} x$. Finally, if either $x,y \in \mods{A}$ or $x,y \in \mods{\neg A}$, then $x \sim_{\Psi \ast A} y$ and  $x \sim_{\Psi' \ast A} y$. 

\item[{\bf (ii)}] {$x \prec_{\Psi} y$ and $x \prec_{\Psi'} y$:} Given that $y \notin \min(\preccurlyeq_{\Psi}, \mods{A})\cup \min(\preccurlyeq_{\Psi'}, \mods{A})$,  we have $x \prec_{\Psi \ast A} y$ and  $x \prec_{\Psi' \ast A} y$.

\item[{\bf (iii)}] {$y \prec_{\Psi} x$ and $y \prec_{\Psi'} x$:} Given that $x \notin \min(\preccurlyeq_{\Psi}, \mods{A})\cup \min(\preccurlyeq_{\Psi'}, \mods{A})$,  we have $y \prec_{\Psi \ast A} x$ and  $y \prec_{\Psi' \ast A} x$.

\end{itemize}
\item[{\bf(c)}] {Regarding natural revision:} Given $x, y \notin \min(\preccurlyeq_{\Psi}, \mods{A})\cup \min(\preccurlyeq_{\Psi'}, \mods{A})$, $x \preccurlyeq_{\Psi} y$ iff $x \preccurlyeq_{\Psi \ast A} y$ and $x \preccurlyeq_{\Psi'}y$ iff $x \preccurlyeq_{\Psi' \ast A} y$. Also: $y \preccurlyeq_{\Psi} x$ iff $y \preccurlyeq_{\Psi \ast A} x$ and $y \preccurlyeq_{\Psi'}x$ iff $y \preccurlyeq_{\Psi' \ast A} x$. 

\end{itemize}
Regarding \IIAInputRev, we have noted, in Proposition \ref{IIAInputandBetas}, that it is equivalent, in the presence of \CRevR{1}--\CRevR{4}, to the conjunction of the  principles \BetaRevR{1} and \BetaRevR{2}.  Proposition 6 of \cite{DBLP:conf/kr/0001C18} establishes that a family of so-called  `POI operators', which includes lexicographic and restrained revision, satisfies a set of principles that are collectively stronger than \BetaRevR{1} and \BetaRevR{2}. However,  if one examines their proof of this claim, one can see that it carries over to a broader  family of {\em BOI} operators, which they mention in their concluding comments and of which all three of our  operators are members. Indeed, the proof makes use of the weaker requirement that $x^+\leq x^-$ employed in the characterisation of BOI operators, rather than the stronger principle $x^+ < x^-$ characteristic of the POI subfamily. 

This completes the proof of Lemma \ref{NRLareElem}. We now show that:

\begin{restatable}{lem}{ElemareNRL}
\label{ElemareNRL}
If an operator is elementary, then it is a lexicographic, restrained or natural revision operator.
\end{restatable}

\noindent \IIAPriorRev~and \NeutralityRev~jointly allows us to represent revision by a given sentence $A$ as a quadruple of functions from prior to posterior relations between two arbitrary worlds $x$ and $y$, such that $x, y\notin\min(\preccurlyeq_{\Psi}, \mods{A})$, one for each of the three following possibilities: (1) $x \in \mods{A}$, $y \in \mods{\neg A}$, (2) $x, y \in \mods{A}$, and (3) $x, y \in \mods{\neg A}$ (the case in which $x \in \mods{\neg A}$, $y \in \mods{A}$ is determined by (1), by virtue of  \NeutralityRev). These functions can be represented by state diagrams in which the set of states is $\{x\prec_{\Psi} y, x\sim_{\Psi} y, y\prec_{\Psi} x\}$ and the edges represent revisions by $A$. For example, $\ast_{\mathrm{L}}$ gives us the following diagram for the function associated with (1):

\bigskip

\begin{centering}
%\begin{tikzpicture}[->,>=stealth',node distance=1.5cm]
%\scalebox{0.6}{%
%%initial node: K
% \node[state] (1) 
% {
% $x\prec y$
%};
%
%
%% Next node: 2
%\node[state,  
%below of=1, 
%] (2)  
%{
% $x\sim y$
% };
% 
% % Next node: 2
%\node[state,  
%below of=2, 
%] (3)  
%{
% $y\prec x$
% };
% 
% 
%
% % draw the paths and and print some Text below/above the graph
% 
%\path (1) edge  [loop right] (1)
%;
%
% \path (2) edge (1)
%;
%
% \path (3) edge [bend right=60]  (1)
%;
%
%}
%\end{tikzpicture}

\includegraphics{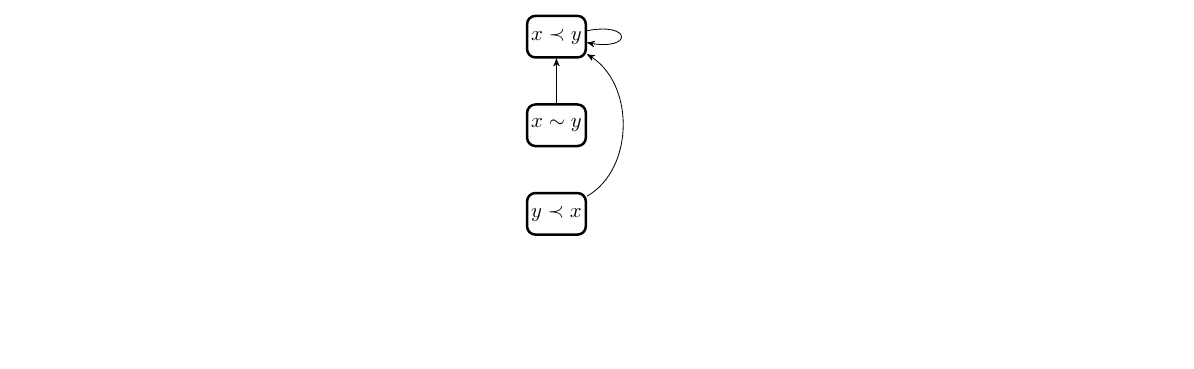}

\end{centering}

%\vspace{-0.75em}
\vspace{-1.75em}

\noindent The postulates \CRevR{1}~and \CRevR{2}~mean that the functions associated with (2) and (3) simply map each state to itself. So whatever degrees of freedom there are, they are associated with (1). Furthermore, the postulates \CRevR{3}~and \CRevR{4}~entail that the arrows in the diagram of the function associated with (1) do not point downwards, given the convention we are adopting for ordering the states vertically. This leaves us with at most {\em six} possible state diagrams:

\bigskip

\begin{centering}
\hspace{0em}
%PICTURE 1
%=====================================================
\hspace{2em}\begin{tikzpicture}[->,>=stealth',node distance=1.5cm]
\scalebox{0.6}{%
%initial node: K
 \node[state] (1) 
 {
 $x\prec y$
};
% Next node: 2
\node[state,  
below of=1, 
] (2)  
{
 $x\sim y$
 };
 % Next node: 2
\node[state,  
below of=2, 
] (3)  
{
 $y\prec x$
 };
 
 \node[  
below of=3, 
] (a)  
{
 {\bf (a)}
 };
 % draw the paths and and print some Text below/above the graph 
\path (1) edge  [loop right] (1)
;
 \path (2) edge (1)
;
 \path (3) edge  [loop right] (3)
;
}
\end{tikzpicture}~\hspace{-2.45em}
%PICTURE 2
%=====================================================
\begin{tikzpicture}[->,>=stealth',node distance=1.5cm]
\scalebox{0.6}{%
%initial node: K
 \node[state] (1) 
 {
 $x\prec y$
};
% Next node: 2
\node[state,  
below of=1, 
] (2)  
{
 $x\sim y$
 };
 % Next node: 2
\node[state,  
below of=2, 
] (3)  
{
 $y\prec x$
 };
 
  \node[  
below of=3, 
] (a)  
{
 {\bf (b)}
 };
 % draw the paths and and print some Text below/above the graph 
\path (1) edge  [loop right] (1)
;
 \path (2) edge (1)
;
 \path (3) edge [bend right=60]  (1)
;
}
\end{tikzpicture}~\hspace{-2.45em}
%PICTURE 3
%=====================================================
\begin{tikzpicture}[->,>=stealth',node distance=1.5cm]
\scalebox{0.6}{%
%initial node: K
 \node[state] (1) 
 {
 $x\prec y$
};
% Next node: 2
\node[state,  
below of=1, 
] (2)  
{
 $x\sim y$
 };
 % Next node: 2
\node[state,  
below of=2, 
] (3)  
{
 $y\prec x$
 };
 
  \node[  
below of=3, 
] (a)  
{
 {\bf (c)}
 };
 % draw the paths and and print some Text below/above the graph 
\path (1) edge  [loop right] (1)
;
 \path (2) edge [loop right] (2)
;
 \path (3) edge [loop right] (3)
;
}
\end{tikzpicture}
\hspace{-2.45em}
%PICTURE 4
%=====================================================
\begin{tikzpicture}[->,>=stealth',node distance=1.5cm]
\scalebox{0.6}{%
%initial node: K
 \node[state] (1) 
 {
 $x\prec y$
};
% Next node: 2
\node[state,  
below of=1, 
] (2)  
{
 $x\sim y$
 };
 % Next node: 2
\node[state,  
below of=2, 
] (3)  
{
 $y\prec x$
 };
 
  \node[  
below of=3, 
] (a)  
{
 {\bf (d)}
 };
 % draw the paths and and print some Text below/above the graph 
\path (1) edge  [loop right] (1)
;
 \path (2) edge [loop right] (2)
;
 \path (3) edge  (2)
;
}
\end{tikzpicture}~\hspace{-2.45em}
%PICTURE 5
%=====================================================
\begin{tikzpicture}[->,>=stealth',node distance=1.5cm]
\scalebox{0.6}{%
%initial node: K
 \node[state] (1) 
 {
 $x\prec y$
};
% Next node: 2
\node[state,  
below of=1, 
] (2)  
{
 $x\sim y$
 };
 % Next node: 2
\node[state,  
below of=2, 
] (3)  
{
 $y\prec x$
 };
 
  \node[  
below of=3, 
] (a)  
{
 {\bf (e)}
 };
 % draw the paths and and print some Text below/above the graph 
\path (1) edge  [loop right] (1)
;
 \path (2) edge (1)
;
 \path (3) edge (2)
;
}
\end{tikzpicture}~\hspace{-2.45em}
%PICTURE 6
%=====================================================
\begin{tikzpicture}[->,>=stealth',node distance=1.5cm]
\scalebox{0.6}{%
%initial node: K
 \node[state] (1) 
 {
 $x\prec y$
};
% Next node: 2
\node[state,  
below of=1, 
] (2)  
{
 $x\sim y$
 };
 % Next node: 2
\node[state,  
below of=2, 
] (3)  
{
 $y\prec x$
 };
 
  \node[  
below of=3, 
] (a)  
{
 {\bf (f)}
 };
 % draw the paths and and print some Text below/above the graph 
\path (1) edge  [loop right] (1)
;
 \path (2) edge  [loop right] (2)
;
 \path (3) edge [bend right=65]  (1)
;
}
\end{tikzpicture}

\end{centering}

\vspace{-4.5em}

\noindent  Diagrams (a), (b) and (c)  respectively correspond to $\ast_{\mathrm{R}}A$, $\ast_{\mathrm{L}}A$  and $\ast_{\mathrm{N}}A$. However, (d) and (e) are inconsistent with \CRevR{2}, on pains of triviality. Indeed assume that there exist two worlds $y, z\in\mods{\neg A}$ and a world $x\in\mods{A}$, such that $z\prec_{\Psi} y\prec_{\Psi} x$. Then $z\sim_{\Psi\ast A} y$, in violation of \CRevR{2}. (f) exhibits a similar inconsistency. Consider  this time the prior TPO given by $y\prec_{\Psi}\{x, w\}$. We have  $z\prec_{\Psi\ast A} y$. Given $y\prec_{\Psi} z$, this is again inconsistent with \CRevR{2}. 

So we have established that \IIAPriorRev~and \NeutralityRev~jointly entail that, for any $A$, $\Psi\ast A$ is equal to one of either $\Psi \ast_{\mathrm{R}}A$,  $\Psi \ast_{\mathrm{L}}A$ or $\Psi \ast_{\mathrm{N}}A$. But it still remains the case that $\ast$ coincides with one elementary operator for one input but with another elementary operator for another, so that, for example, $\Psi\ast A= \Psi \ast_{\mathrm{R}}A$ while $\Psi\ast A= \Psi \ast_{\mathrm{R}}A$. This is ruled out by the final condition \IIAPriorRev. \qed
\end{pproof}

\vspace{1em}

\iLITwoTriv*

\begin{pproof} Assume $\cbel{\Psi \ast A} = \mathrm{Cn}(\cbel{\Psi \contract \neg A} \cup \{A\})$. By Success,  $A\in\bel{\Psi\ast A}$. By the Ramsey Test, $\top\Rightarrow A\in \cbel{\Psi\ast A}$ and so $\top\Rightarrow A\in \mathrm{Cn}(\cbel{\Psi \contract \neg A} \cup \{A\})$. But then, since $\top\Rightarrow A\notin L$ and, as we have stipulated, for $\Delta\subseteq L_c$, $\Cn(\Delta)=\Delta\cup \Cn(\Delta\cap L)$, it must be the case that $\top\Rightarrow A\in \cbel{\Psi \contract \neg A}$. Hence, by the Ramsey Test again, it follows that $A \in \bel{\Psi\contract \neg A}$. From \KCon{2}, we then have $A\in \bel{\Psi}$. By a similar chain of reasoning, we can establish that $\neg A\in \bel{\Psi}$. \qed
\end{pproof}

\vspace{1em}

\ChopratoDP*

\begin{pproof}

\begin{itemize}
\item[{\bf (a)}] {$i=1$:} Assume that $x, y\in \mods{A}$. We want to show that $x \preccurlyeq_{(\Psi \contract \neg A) + A} y$ iff $x \preccurlyeq_{\Psi} y$. Since $\contract$ satisfies \CConR{1}, we have $x \preccurlyeq_{\Psi \contract \neg A} y$ iff $x \preccurlyeq_{\Psi} y$. Since $+$ satisfies \CRevR{1}, we have $x \preccurlyeq_{(\Psi \contract \neg A) + A} y$ iff $x \preccurlyeq_{\Psi \contract \neg A} y$, and we are done.
\item[{\bf (b)}]  {$i=2$:} Analogous to $i=1$. 
\item[{\bf (c)}] {$i=3$:} Assume that $x\in \mods{A}$, $y\in \mods{\neg A}$. We want to show that, if $x \prec_{\Psi} y$, then $x \prec_{(\Psi \contract \neg A) + A} y$. Assume $x \prec_{\Psi} y$. By \CConR{3},  $x \prec_{\Psi \contract \neg A} y$. By \CRevR{3}, which + satisfies, $x \preccurlyeq_{(\Psi \contract \neg A) + A} y$, as required.
\item[{\bf (d)}] {$i=4$:} Analogous to $i=3$. \qed
\end{itemize}
\end{pproof}

\vspace{1em}

\AltNecSuffNLIcopy*

\begin{pproof} We prove each direction of the claim in the form of a separate lemma. Regarding the right-to-left direction:

\begin{restatable}{lem}{SuffNLI}
\label{SuffNLI}
If $\ast$ is an elementary revision operator, $\contract$ satisfies \CConR{1}-\CConR{4},  and $\contract$ and $\ast$ satisfy \CConRevR{1}-\CConRevR{4}, then $\ast$ and $\contract$ satisfy \NLI. 
\end{restatable}

\noindent We want to show that, given the relevant assumptions, $x \preccurlyeq_{\Psi\ast A} y$ iff $x \preccurlyeq_{(\Psi \contract \neg A)\ast A} y$. We consider two cases: 
\begin{itemize}

\item[{\bf (1)}] {$x$ or $y\in\min(\preccurlyeq_{\Psi}, \mods{A})$:} We first show that, if $y\in\min(\preccurlyeq_{\Psi}, \mods{A})$, then $x \preccurlyeq_{\Psi\ast A} y$ iff $x \preccurlyeq_{(\Psi \contract \neg A)\ast A} y$:
\begin{itemize}

\item[{\bf (a)}] {Left-to-right direction ($x \preccurlyeq_{\Psi\ast A} y\Rightarrow x \preccurlyeq_{(\Psi \contract \neg A)\ast A} y$):} By Success, $\min(\preccurlyeq_{\Psi}, \mods{A}) = \min(\preccurlyeq_{\Psi\ast A}, W)$, so it follows from $y\in\min(\preccurlyeq_{\Psi}, \mods{A})$ that $y\in \min(\preccurlyeq_{\Psi\ast A}, W)$. Assume $x \preccurlyeq_{\Psi\ast A} y$. Then, since $y\in \min(\preccurlyeq_{\Psi\ast A}, W)$, we have $x\in \min(\preccurlyeq_{\Psi\ast A}, W)$ and hence $x\in\min(\preccurlyeq_{\Psi}, \mods{A})$. From this, it then follows by \CConR{2} that $x\in\min(\preccurlyeq_{\Psi \contract \neg A},\mods{A})$. Finally, by \CRevR{1}, we recover $x\in\min(\preccurlyeq_{(\Psi \contract \neg A)\ast A},\mods{A})$, from which we have, by Success, $x\in\min(\preccurlyeq_{(\Psi \contract \neg A)\ast A},W)$ and so $x \preccurlyeq_{(\Psi \contract \neg A)\ast A} y$, as required.

\item[{\bf (b)}] {Right-to-left direction ($x \preccurlyeq_{(\Psi \contract \neg A)\ast A} y \Rightarrow x \preccurlyeq_{\Psi\ast A} y$):}  It follows from \CConR{2} that $\min(\preccurlyeq_{\Psi}, \mods{A}) = \min(\preccurlyeq_{\Psi \contract \neg A},\mods{A})$. From Success, we also have $\min(\preccurlyeq_{\Psi \contract \neg A},\mods{A}) = \min(\preccurlyeq_{(\Psi \contract \neg A)\ast A},W)$. So it follows from $y\in\min(\preccurlyeq_{\Psi}, \mods{A})$ that $y\in\min(\preccurlyeq_{(\Psi \contract \neg A)\ast A},W)$. Assume $x \preccurlyeq_{(\Psi \contract \neg A)\ast A} y$. Then, since $y\in\min(\preccurlyeq_{(\Psi \contract \neg A)\ast A},W)$, we have  $x\in\min(\preccurlyeq_{(\Psi \contract \neg A)\ast A},W)$, and hence $x\in\min(\preccurlyeq_{\Psi}, \mods{A})$. But by Success, $\min(\preccurlyeq_{\Psi}, \mods{A}) = \min(\preccurlyeq_{\Psi\ast A}, W)$. Hence $x\in \min(\preccurlyeq_{\Psi\ast A}, W)$ and and so $x \preccurlyeq_{\Psi\ast A} y$, as required.

\end{itemize}
The same equivalence can be even more immediately established for the case in which $x\in\min(\preccurlyeq_{\Psi}, \mods{A})$:
\begin{itemize}

\item[{\bf (a)}] {Left-to-right direction ($x \preccurlyeq_{\Psi\ast A} y\Rightarrow x \preccurlyeq_{(\Psi \contract \neg A)\ast A} y$): } 
%By Success, $\min(\preccurlyeq_{\Psi}, \mods{A}) = \min(\preccurlyeq_{\Psi\ast A}, W)$, so it follows from $x\in\min(\preccurlyeq_{\Psi}, \mods{A})$ that $x\in \min(\preccurlyeq_{\Psi\ast A}, W)$. \jake{do we even need what precedes?} 
From $x\in\min(\preccurlyeq_{\Psi}, \mods{A})$ and  \CConR{2}, it follows that $x\in\min(\preccurlyeq_{\Psi \contract \neg A},\mods{A})$. From this, by \CRevR{1}, we recover $x\in\min(\preccurlyeq_{(\Psi \contract \neg A)\ast A},\mods{A})$, from which we have, by Success, $x\in\min(\preccurlyeq_{(\Psi \contract \neg A)\ast A},W)$ and so $x \preccurlyeq_{(\Psi \contract \neg A)\ast A} y$, as required.

\item[{\bf (b)}] {Right-to-left direction ($x \preccurlyeq_{(\Psi \contract \neg A)\ast A} y \Rightarrow x \preccurlyeq_{\Psi\ast A} y$):}  By Success, $\min(\preccurlyeq_{\Psi}, \mods{A}) = \min(\preccurlyeq_{\Psi\ast A}, W)$. Hence, since $x\in\min(\preccurlyeq_{\Psi}, \mods{A})$, we have $x\in \min(\preccurlyeq_{\Psi\ast A}, W)$ and and so $x \preccurlyeq_{\Psi\ast A} y$, as required.

\end{itemize} 

\item[{\bf (2)}] {$x, y\notin\min(\preccurlyeq_{\Psi}, \mods{A})$:} If $x, y\in\mods{A}$ or $x,y\in\mods{\neg A}$, the required result is immediate, following from either \CRevR{1} and \CConR{2} or  \CRevR{2} and \CConR{1}. This leaves the two cases in which $x$ and $y$ differ with respect to their membership of $\mods{A}$. These are dealt with in the same manner, so we shall simply establish the result for the case in which $x\in\mods{A}$ and $y\in\mods{\neg A}$. Below, we find the three possible state diagrams for the case in which $x\in\mods{A}$ and $y\in\mods{\neg A}$, and $x, y\notin\min(\preccurlyeq_{\Psi}, \mods{A})$, where the solid arrows denote transitions by revision by $A$. To each of these, we have added dashed arrows to denote permissible transitions by contraction by $\neg A$. Note that, since we do not assume that $\contract$ satisfies the obvious analogue for contraction of \IIAPriorRev, multiple dashed arrows from each state are permitted. Postulates \CConR{3} and \CConR{4} ensure that these dashed arrows do not take us downwards.  Postulates \CConRevR{3} and \CConRevR{4} ensure that they do not take us further up than the solid arrow that originates in the same state.

\bigskip

\begin{centering}
%PICTURE 1
%=====================================================
\begin{tikzpicture}[->,>=stealth',node distance=1.5cm]
\scalebox{0.6}{%
%initial node: K
 \node[state] (1) 
 {
 $x\prec y$
};
% Next node: 2
\node[state,  
below of=1, 
] (2)  
{
 $x\sim y$
 };
 % Next node: 2
\node[state,  
below of=2, 
] (3)  
{
 $y\prec x$
 };
 
 \node[  
below of=3, 
] (a)  
{
 {\bf (a)}
 };
 % draw the paths and and print some Text below/above the graph 
 %solid paths
\path (1) edge  [loop right] (1)
;
 \path (2) edge [transform canvas={xshift=1.5mm}] (1)
;
 \path (3) edge  [loop right] (3)
;
 %dashed paths
 \path (1) edge  [style=dashed, loop left] (1)
;
 \path (2) edge  [style=dashed, loop left] (2)
;
 \path (2) edge  [style=dashed, transform canvas={xshift=-1.5mm}] (1)
;
 \path (3) edge  [style=dashed, loop left] (3)
;
}
\end{tikzpicture}\hspace{-2em}
%PICTURE 2
%=====================================================
\begin{tikzpicture}[->,>=stealth',node distance=1.5cm]
\scalebox{0.6}{%
%initial node: K
 \node[state] (1) 
 {
 $x\prec y$
};
% Next node: 2
\node[state,  
below of=1, 
] (2)  
{
 $x\sim y$
 };
 % Next node: 2
\node[state,  
below of=2, 
] (3)  
{
 $y\prec x$
 };
 
  \node[  
below of=3, 
] (a)  
{
 {\bf (b)}
 };
 % draw the paths and and print some Text below/above the graph 
 %solid paths
\path (1) edge  [loop right] (1)
;
 \path (2) edge [transform canvas={xshift=1.5mm}] (1)
;
 \path (3) edge [bend right=60]  (1)
;
 %dashed paths
 \path (1) edge  [style=dashed, loop left] (1)
;
 \path (2) edge  [style=dashed, loop left] (2)
;
 \path (2) edge  [style=dashed, transform canvas={xshift=-1.5mm}] (1)
;
 \path (3) edge  [style=dashed, loop left] (3)
;
 \path (3) edge  [style=dashed] (2)
;
 \path (3) edge  [style=dashed, bend left=65] (1)
;
}
\end{tikzpicture}
%\hspace{-2em}
%PICTURE 3
%=====================================================
\begin{tikzpicture}[->,>=stealth',node distance=1.5cm]
\scalebox{0.6}{%
%initial node: K
 \node[state] (1) 
 {
 $x\prec y$
};
% Next node: 2
\node[state,  
below of=1, 
] (2)  
{
 $x\sim y$
 };
 % Next node: 2
\node[state,  
below of=2, 
] (3)  
{
 $y\prec x$
 };
 
  \node[  
below of=3, 
] (a)  
{
 {\bf (c)}
 };
 % draw the paths and and print some Text below/above the graph 
 %solid paths
\path (1) edge  [loop right] (1)
;
 \path (2) edge [loop right] (2)
;
 \path (3) edge [loop right] (3)
;
 %dashed paths
 \path (1) edge  [style=dashed, loop left] (1)
;
 \path (2) edge  [style=dashed, loop left] (2)
;
 \path (3) edge  [style=dashed, loop left] (3)
;
}
\end{tikzpicture}

\end{centering}

\vspace{-4em}

\noindent \NLI~is then satisfied iff any trip along a dashed arrow $d$ and then a solid arrow $s$ ends in the state which is pointed to by the solid arrow $s'$ that originates in the same place as $d$. This is easily verified to be true.

\end{itemize}

\noindent Regarding the  left-to-right direction, we will actually prove the following stronger claim:

\begin{restatable}{lem}{ConRevNec}
\label{ConRevNec}
If $\ast$ satisfies \CRevR{1}--\CRevR{4}, then there exists $\contract$ such that $\ast$ and $\contract$ satisfy \NLI~only if $\ast$ and $\contract$ satisfy \CConRevR{1}--\CConRevR{4}
\end{restatable}

\noindent We establish the necessity of each of \CConRevR{1}--\CConRevR{4} in turn:
\begin{itemize}
\item[{\bf (1)}] {Regarding \CConRevR{1}:} Assume that \CConRevR{1} fails, so that $x, y\in\mods{A}$ and either (i) $x\preccurlyeq_{\Psi \contract \neg A} y$ but $y\prec_{\Psi\ast A} x$ or (ii) $y\prec_{\Psi \contract \neg A} x$ but $x\preccurlyeq_{\Psi\ast A} y$. Assume (i). From $x\preccurlyeq_{\Psi \contract \neg A} y$, it follows, by \CRevR{1}, that $x \preccurlyeq_{(\Psi\contract \neg A) \ast A} y$. By \NLI, we then have $x \preccurlyeq_{\Psi\ast A} y$. Contradiction. Assuming (ii) leads to a contradiction in an analogous manner.

\item[{\bf (2)}] {Regarding \CConRevR{2}:} As for \CConRevR{1}, using  \CRevR{2}.

\item[{\bf (3)}]  {\sloppy {Regarding \CConRevR{3}:} Assume  \CConRevR{3} fails, so that $x\in\mods{A}$, $y\in\mods{\neg A}$,  $x\prec_{\Psi \contract \neg A} y$, but $y\preccurlyeq_{\Psi\ast A} x$. From $x\prec_{\Psi \contract \neg A} y$, by \CRevR{3}, it follows that  $x\prec_{(\Psi\contract \neg A) \ast A} y$. By \NLI, we then have $x \prec_{\Psi\ast A} y$. Contradiction.}

\item[{\bf (4)}] {Regarding \CConRevR{4}:} As for \CConRevR{3}, using  \CRevR{4}.\qed
\end{itemize}
\end{pproof}

\vspace{1em}

%=====================================================

\SWPUandConRev*

\begin{pproof} We prove the result in two parts. Firstly we establish the following strengthening of the right-to-left direction of the claim:

\begin{restatable}{lem}{TQConRev}
\label{TQConRev}
Given \SPU~and \WPU, for all $1\leq i\leq 4$, \CRevR{i} entails \CConRevR{i}, where: 
\begin{tabbing}
\=BLAHBLI: \=\kill
\> \CConRevR{1} \>   If $x, y\in\mods{A}$, then $x\preccurlyeq_{\Psi \contract \neg A} y$ iff $x\preccurlyeq_{\Psi\ast A} y$ \\
\> \CConRevR{2} \>    If $x, y\in\mods{\neg A}$, then $x\preccurlyeq_{\Psi \contract \neg A} y$ iff $x\preccurlyeq_{\Psi\ast A} y$ \\
\> \CConRevR{3} \>     If $x\in\mods{A}$, $y\in\mods{\neg A}$, and $x\prec_{\Psi \contract \neg A} y$,then  $x\prec_{\Psi\ast A} y$.  \\
\> \CConRevR{4} \>     If $x\in\mods{A}$, $y\in\mods{\neg A}$, and $x\preccurlyeq_{\Psi \contract \neg A} y$,  then  $x\preccurlyeq_{\Psi\ast A} y$.   \\[-0.25em]
\end{tabbing} 
\vspace{-0.75em}

\end{restatable}

\noindent 

\begin{itemize}

\item[{\bf (a)}] {Regarding $i=1, 2$:} We provide the proof for $i=1$, since the case in which $i=2$ is handled analogously. Assume $x, y \in \mods{A}$. From left to right: Assume $x \preccurlyeq_{\Psi \contract \neg A} y$. By the contrapositve of \SPU, either $x \preccurlyeq_{\Psi} y$ or $x \preccurlyeq_{\Psi \ast A} y$. If the latter holds, we are done. So assume that $x \preccurlyeq_{\Psi} y$. Then the required result follows by \CRevR{1}. From right to left:  Assume $x \preccurlyeq_{\Psi \ast A} y$. By  \CRevR{1}, $x \preccurlyeq_{\Psi} y$. By \WPU, $x \preccurlyeq_{\Psi \contract \neg A} y$, as required.

\item[{\bf (b)}] {Regarding $i=3, 4$:} We provide the proof for $i=3$, since the case in which $i=4$ is handled analogously (using \SPU~rather than \WPU). We derive the contrapositive. Assume $x \in \mods{A}$, $y \in \mods{\neg A}$ and $y \preccurlyeq_{\Psi\ast A} x$. If $y \preccurlyeq_{\Psi} x$, then, from $y \preccurlyeq_{\Psi\ast A} x$, we have $y \preccurlyeq_{\Psi \contract \neg A} x$, by \WPU, as required. So assume $x \prec_{\Psi} y$. By \CRevR{3}, $x \prec_{\Psi\ast A} y$. Contradiction.

\end{itemize}
This completes the proof of Lemma \ref{TQConRev}. Concerning the left-to-right direction of our principal claim, we show:
{\sloppy
\begin{restatable}{lem}{ConRevTQ}
\label{ConRevTQ}
Given \CConR{3}~and \CConR{4},  \CConRevR{1}~to \CConRevR{4} collectively entail both \SPU\,  and \WPU. (Alternatively: Given \CConR{1}~to \CConR{4}, \CConRevR{3}~and \CConRevR{4} jointly entail both \SPU~and \WPU.)
\end{restatable}
}

\noindent We just prove this in relation to \WPU, using \CConRevR{1}, \CConRevR{2}, \CConRevR{3} and \CConR{4}. The proof in relation to \SPU~is analogous but uses \CConRevR{1}, \CConRevR{2}, \CConRevR{4} and \CConR{3}~instead. Assume that $x\preccurlyeq_{\Psi} y$ and $x\preccurlyeq_{\Psi \ast A} y$. We want to show $x\preccurlyeq_{\Psi \contract\neg A} y$. If (a) $x, y\in\mods{A}$, (b) $x, y\in\mods{\neg A}$, or (c) $x \in\mods{\neg A}$ and $y\in\mods{A}$, this follows from $x\preccurlyeq_{\Psi \ast A} y$, by \CConRevR{1}, \CConRevR{2}~or \CConRevR{3}, respectively. If (d) $x \in\mods{A}$ and $y\in\mods{\neg A}$, then it follows from $x\preccurlyeq_{\Psi} y$, by \CConR{4}.

Note that we can also substitute \CConR{1} and \CConR{2} for \CConRevR{1} and  \CConRevR{2}, obtaining the required result from $x\preccurlyeq_{\Psi} y$ instead of $x\preccurlyeq_{\Psi \ast A} y$.\qed
\end{pproof}

\vspace{1em}

\LeviIDRatNat*

\begin{pproof}
We prove the claim by establishing that \iLIR~ensures that $\preccurlyeq_{\Psi \ast A}$ is the `flattest' TPO--in a technical sense to be defined below--such that the following lower bound constraint is satisfied:

\begin{tabbing}
\=BLAHBLI: \=\kill
\> \> $ \cbel{\Psi \contract \neg A} \cup \{A\}\subseteq \cbel{\Psi \ast A}$ \\[-0.25em]
\end{tabbing} 
\vspace{-0.75em}

\noindent In view of Definitions 20 and 21 of \cite{lehmann1992does}, the upshot of this is then that $\preccurlyeq_{(\Psi \contract \neg A)\astN A}$ is the unique TPO corresponding to the rational closure of $\cbel{\Psi \contract \neg A} \cup \{A\}$.

We first note that, given AGM, the lower bound principle can be semantically expressed as follows:

\begin{tabbing}
\=BLAHBLI: \=\kill
\> \> (a) If $x \prec_{\Psi \contract \neg A} y$, then $x \prec_{\Psi \ast A} y$ and \\
\> \> (b) $\min(\preccurlyeq_{\Psi \ast A}, W)\subseteq \mods{A}$ \\[-0.25em]
\end{tabbing} 
\vspace{-0.75em}

\noindent Indeed, the lower bound constraint simply amounts to the conjunction of Success, which is equivalent to (b), with the claim that $ \cbel{\Psi \contract \neg A}\subseteq \cbel{\Psi \ast A}$, which is equivalent to (a). With this in hand, we now prove two lemmas. First: 

\begin{restatable}{lem}{SPP}
\label{SPP}
If $\ast$ and $\contract$ satisfy \iLIRSem, then they satisfy the lower bound principle.
\end{restatable}

\noindent Establishing satisfaction of (b) is trivial. So we just need to establish satisfaction of (a). Assume $x\prec_{\Psi \contract \neg A}y$. Given \iLIR, we will have $x~\prec_{\Psi \ast A}~y$ iff either

\begin{itemize}

\item[(1)] $x \in \min(\preccurlyeq_{\Psi \contract \neg A}, \mods{A})$ and $y \notin \min(\preccurlyeq_{\Psi \contract \neg A}, \mods{A})$, or 

\item[(2)] $x, y \notin \min(\preccurlyeq_{\Psi \contract \neg A}, \mods{A})$ and $x\prec_{\Psi \contract \neg A}y$

\end{itemize}

\noindent Note first that $y \notin \min(\preccurlyeq_{\Psi \contract \neg A}, \mods{A})$. Indeed, assume that this were false. Since we know that $\min(\preccurlyeq_{\Psi \contract \neg A}, \mods{A}) \subseteq \min(\preccurlyeq_{\Psi \contract \neg A}, W)$, this would mean that $y\in \min(\preccurlyeq_{\Psi \contract \neg A}, W)$. But this is inconsistent with $x\prec_{\Psi \contract \neg A}y$. This leaves us with two possibilities. The first is that $x, y \notin \min(\preccurlyeq_{\Psi \contract \neg A}, \mods{A})$, which, given $x\prec_{\Psi \contract \neg A}y$,  places us in case (2). The second is that $x \in \min(\preccurlyeq_{\Psi \contract \neg A}, \mods{A})$ and $y \notin \min(\preccurlyeq_{\Psi \contract \neg A}, \mods{A})$, which places us in case (1). Either way, then, $x~\prec_{\Psi \ast A}~y$, as required. This completes the proof of Lemma \ref{SPP}.

For our second lemma, we will make use of the convenient representation of TPOs by their corresponding {\em ordered partitions} of $W$. The ordered partition $\langle S_1, S_2, \ldots S_m\rangle$ of $W$ corresponding to a TPO $\preccurlyeq$ is such that $x \preccurlyeq y$ iff $r(x, \preccurlyeq) \leq$ \mbox{$r(y, \preccurlyeq)$,} where $r(x, \preccurlyeq)$ denotes the `rank' of $x$ with respect to $\preccurlyeq$ and is defined by taking $S_{r(x, \preccurlyeq)}$ to be the cell in the partition that contains $x$. 

This lemma is given as follows:

\begin{restatable}{lem}{FlattestLI}
\label{FlattestLI}
$\preccurlyeq_{(\Psi \contract \neg A) \astN A}~\sqsupseteq~~\preccurlyeq$, for any TPO $\preccurlyeq$ satisfying the lower bound principle. 
\end{restatable}

\noindent where:

{\sloppy 
\begin{definition}
\label{dfn:Flatter}
$\sqsupseteq$ is a binary relation on the set of TPOs over $W$ such such that, for any TPOs $\preccurlyeq_1$ and $\preccurlyeq_2$, whose corresponding ordered partitions are given by $\langle S_1, S_2, \ldots, S_m \rangle$ and $\langle T_1, T_2, \ldots, T_m \rangle$ respectively,   
$
\preccurlyeq_1~
\sqsupseteq~
 \preccurlyeq_2
$
iff
either (i) $S_i = T_i$ for all $i = 1, \ldots, m$, or (ii) $S_i \supset T_i$ for the first $i$ such that $S_i \neq T_i$. 
\end{definition}
}

\noindent $\sqsupseteq$ partially orders $TPO(W)$ according to  comparative `flatness', with the flatter TPOs appearing `greater' in the ordering, so that $\preccurlyeq_1~\sqsupseteq~\preccurlyeq_2$ iff $\preccurlyeq_1 $ is at least as as flat as $\preccurlyeq_2$.

Let $\langle T_1,\ldots, T_m\rangle$ be the ordered partition corresponding to the TPO $\preccurlyeq_{(\Psi \contract \neg A) \astN A}$, which we will denote by $ \preccurlyeq_{\mathbb{N}}$. Let $\preccurlyeq $ be any TPO satisfying the lower bound condition: 
\begin{itemize}

\item[(a)] If $x \prec_{\Psi \contract \neg A} y$, then $x \prec y$ and 

\item[(b)] $\min(\preccurlyeq, W)\subseteq \mods{A}$. 

\end{itemize}
Let $\langle S_1,\ldots, S_n\rangle$ be its corresponding ordered partition. We must show that the following relation holds:  $\preccurlyeq_{\mathbb{N}}~\sqsupseteq~\preccurlyeq$.

If $T_i=S_i$ for all $i$, then we are done. So let $i$ be minimal such that $T_i\neq S_i$. We must show $S_i\subset T_i$. So let $y\in S_i$ and assume, for contradiction, that  $y\notin T_i$. We know that $T_i\neq \varnothing$, since, otherwise, $\bigcup_{j < i} T_j = W$, hence $\bigcup_{j < i} S_j = W$ and so $S_i=\varnothing$, contradicting $S_i\neq T_i$. So let $x\in T_i$. Then, since $y\notin T_i$, we have $x \prec_{\mathbb{N}} y$. We are going to show that this entails that $\exists z$ such that 
\begin{itemize}

\item[(i)] $z\sim_{\mathbb{N}} x$, i.e.~$z\in T_i$ and 

\item[(ii)]  $z\prec y$. 

\end{itemize}
But if it were the case that $z\prec y$, then, since $y\in S_i$, $z\in \bigcup_{j < i} S_j = \bigcup_{j < i} T_j$, contradicting $z\in T_i$. Hence $y\in T_i$ and so we can conclude that $S_i\subset T_i$, as required.

Given $x \prec_{\mathbb{N}} y$, by the definition of $\astN$, one of the following must hold:
\begin{itemize}

\item[{\bf (1)}] {$x \in \min(\preccurlyeq_{\Psi \contract \neg A}, \mods{A})$ and $y \notin \min(\preccurlyeq_{\Psi \contract \neg A}, \mods{A})$}:

\begin{itemize}
\item[{\bf (a)}] $y\in \mods{A}$: We again have $x\prec_{\Psi \contract \neg A} y$, and so $x\prec y$ once more. Then, $x$ plays the role of the required $z$, satisfying conditions (i) and (ii) above, and we are done.

\item[{\bf (b)}] $y\in \mods{\neg A}$: If $x \in \min(\preccurlyeq, W)$, then by (b) and $y\in \mods{\neg A}$, it follows that $x\prec y$ and we are done. So assume $x \notin \min(\preccurlyeq, W)$. Let $z\in\min(\preccurlyeq,W)$. We have $z\prec x$ and, by (b), $z\in\mods{A}$. By the contrapositive of (a), it follows from $z\prec x$ that $z\preccurlyeq_{\Psi \contract \neg A} x$. From this, $x \in \min(\preccurlyeq_{\Psi \contract \neg A}, \mods{A})$ and $z\in\mods{A}$, we then have $z \in \min(\preccurlyeq_{\Psi \contract \neg A}, \mods{A})$. But, by the definition of $\astN$, $x, z \in \min(\preccurlyeq_{\Psi \contract \neg A}, \mods{A})$ entails that   $z \sim_{\mathbb{N}} x$. Finally, since $z\in\min(\preccurlyeq,W)$, $\min(\preccurlyeq,W)\subseteq\mods{A}$ and $y\in \mods{\neg A}$, we have  $z \prec y$. Here, $z$ satisfies conditions (i) and (ii) above and we are done. 
\end{itemize}

\item[{\bf (2)}] {$x, y \notin \min(\preccurlyeq_{\Psi \contract \neg A}, \mods{A})$ and $x\prec_{\Psi \contract \neg A} y$:} Since  $x\prec_{\Psi \contract \neg A} y$, by (a) above, we have $x\prec y$.

\end{itemize}
This completes the proof of Lemma \ref{FlattestLI}. \qed
\end{pproof}

\vspace{1em}

%=====================================================

\ClosureImp*

\begin{pproof}
%We first recall the definition of \iHIRSem, from \jake{ref}:
%
%\begin{tabbing}
%BLI\=BLAHBLI: \=\kill
%\>  \iHIRSem  \> $\preccurlyeq^\contract_{A} = \preccurlyeq\STQ \preccurlyeq^\ast_{\neg A}$ \\[-0.25em]
%\end{tabbing} 
%\vspace{-1em}
%
%\noindent where $\STQ$ is a binary TPO combination operator, defined as follows: 
%
%\begin{definition}
%$\STQ$ is a function returning, from any pair of TPOs $\preccurlyeq_1$ and $\preccurlyeq_2$, a TPO $\preccurlyeq_1\STQ\preccurlyeq_2$ whose corresponding ordered partition $\langle T_1, T_2, \ldots, T_m\rangle$ of $W$ is constructed inductively as follows:
%\[
%T_{i} = \bigcup_{j \in \{1,2\}} \min(\preceq_j, \bigcap_{k<i}T_{k}^c)
%\]
%where `$T^c$' denotes the complement of set $T$
%and $m$ is minimal such that $\bigcup_{i\leq m} T_i = W$. 
%\end{definition}
%
%\noindent where the ordered partition $\langle S_1, S_2, \ldots S_m\rangle$ of $W$ corresponding to a TPO $\preceq_{\Psi}$ is such that $x \preceq_{\Psi} y$ iff $r(x, \preceq_{\Psi}) \leq$ \mbox{$r(y, \preceq_{\Psi})$,} where $r(x, \preceq_{\Psi})$ denotes the `rank' of $x$ with respect to $\preceq_{\Psi}$ and is defined by taking $S_{r(x, \preceq_{\Psi})}$ to be the cell in the partition that contains $x$. 
%
Assume for reductio that there is a closure operator $\C$, satisfying \RID~and such that both \NLI~and $\cbel{\Psi \ast A} = \mathrm{C}(\cbel{\Psi \contract \neg A} \cup \{A\})$ are true. We will show that the following then holds: If $A\in\bel{\Psi}$, then $\cbel{\Psi\contract\neg A}=\cbel{\Psi\ast A}$.

Assume $A\in\bel{\Psi}$. By the AGM postulate \CRevS{3}, which entails that, if $\neg A\notin [\Psi]$, then $[\Psi\contract \neg A]=[\Psi]$, it then follows that $A\in\bel{\Psi\contract \neg A}$. Hence, $\C(\cbel{\Psi\contract \neg A}\cup\{A\})= \C(\cbel{\Psi\contract \neg A})$. Since $\cbel{\Psi\contract \neg A}$ is rational, by \RID, $\C(\cbel{\Psi\contract \neg A}) = \cbel{\Psi\contract \neg A}$. Hence, $\C(\cbel{\Psi\contract \neg A}\cup\{A\})= \cbel{\Psi\contract \neg A}$. Given AGM and the Ramsey Test, \NLI~is equivalent to $\cbel{\Psi\ast A}=\cbel{(\Psi\contract\neg A)\ast A}$. So, in view of the previous equality, we can conclude that $\cbel{\Psi\ast A}=\cbel{\Psi\contract \neg A}$.

But now, the following model provides a case in which $A\in\bel{\Psi}$, but $\cbel{\Psi\contract\neg A}\neq\cbel{\Psi\ast A}$: 

\begin{centering}

\begin{tikzpicture}[->,>=stealth']
\scalebox{0.9}{%
\node[state,
 text width=2.75cm] (K)   
 {
\begin{tabular}[c]{c|c|c|c}
     \multicolumn{1}{c|}{$AB$} &  $A\overline{B}$ & $\overline{A}B$    &\multicolumn{1}{c}{$\overline{A}\overline{B}$} \\
    \hline
        & & & $z$   \\
        & $x$  & $y$  &   \\
      $w$   &   &    &  \\[0.25em]
  \end{tabular}
};

% Next node: astR
 \node[state,       % layout (defined above)
 node distance=3.5cm,     % distance to K
 text width=2.75cm,        % max text width
below of=K,        % Position is to the right of K
 yshift=+0cm] (astR)    % move 3cm in y
 {%                     % posistion relative to the center of the 'box'
 
\begin{tabular}[c]{c|c|c|c}
     \multicolumn{1}{c|}{$AB$} &  $A\overline{B}$ & $\overline{A}B$    &\multicolumn{1}{c}{$\overline{A}\overline{B}$} \\
    \hline
         &   &    & $z$  \\ 
    &   & $y$  &   \\    
            & $x$  &  &   \\
      $w$  & & &   \\[0.25em]
  \end{tabular}
% 
% 
%\begin{tabular}[c]{c|c||c}
%     \multicolumn{1}{c|}{$A$} &  $\neg A$ &\multicolumn{1}{c}{$\kappa$} \\
%    \hline
%    \rule{0pt}{2em}
%      $x$  &    & 2 \\
%              & $y$~$z$ & 1 \\
%       $w$  &       & 0 \\[0.25em]
%  \end{tabular}
};

 % draw the paths and and print some Text below/above the graph
 \path (K) edge  node[anchor=east, right]
                   {
                   $\ast_{\mathrm{R/L}}A$
                   } (astR)
;

 \path (K)  	edge[loop left]    node[anchor=east, right]{$\contract_{\mathrm{STQL}}\neg A$} (K)
;

}
\end{tikzpicture}

\end{centering}
\qed
\end{pproof}

\hfill

\end{document}